\documentclass[10pt,twocolumn,letterpaper]{article}
\usepackage[margin=0.75in,columnsep=0.25in]{geometry}
\usepackage[T1]{fontenc}
\usepackage{times}
\usepackage{cite}
\usepackage{amsmath}
\usepackage{graphicx}
\usepackage{textcomp}
\usepackage{booktabs}
\usepackage{xspace}
\usepackage{xparse}
\usepackage{url}

\usepackage[hidelinks]{hyperref}
\gdef\xfigwd{0pt}
\usepackage{etoolbox}
\makeatletter\def\ax@tab{table}\g@addto@macro\@floatboxreset{\ifx\@captype\ax@tab\footnotesize\fi}\makeatother
\usepackage[font=small,labelfont=bf]{caption}

\newif\ifdraftpending \draftpendingfalse

\newcommand{\NSized}{204}

\newcommand{\FirstAboveSixtyEightParams}{395.9M}

\newcommand{\NGroups}{54}
\newcommand{\NModels}{208}
\newcommand{\NSmall}{24}
\newcommand{\NSmallCrivo}{5}
\newcommand{\BoardMaxFull}{71.24}
\newcommand{\MapSSlope}{1.012}
\newcommand{\MapSIntercept}{\ensuremath{-}1.05}
\newcommand{\MapSMaxDev}{0.009}
\newcommand{\MapOSlope}{1.217}
\newcommand{\MapOIntercept}{1.33}
\newcommand{\MapOMaxDev}{0.010}
\newcommand{\MapN}{208}
\newcommand{\FullSeventeen}{11.11}
\newcommand{\SSSeventeen}{15.88}

\newcommand{\FullThirtyTwo}{13.70}

\newcommand{\FullSixtyEight}{17.69}

\newcommand{\SSOneFifty}{24.93}

\newcommand{\FullOneB}{28.96}

\newcommand{\FullPrumo}{12.21}
\newcommand{\SSPrumo}{15.78}
\newcommand{\NDPrumo}{5.60}

\newcommand{\RPubSS}{0.981}
\newcommand{\RPubSSci}{\mbox{[0.974, 0.986]}}
\newcommand{\RhoPubSS}{0.978}
\newcommand{\RhoPubSSci}{\mbox{[0.969, 0.983]}}
\newcommand{\RPartial}{0.965}
\newcommand{\RPartialci}{\mbox{[0.951, 0.974]}}
\newcommand{\RBucketMin}{0.94}
\newcommand{\RBucketMax}{0.98}
\newcommand{\NBuckets}{5}
\newcommand{\VOneTwoDiff}{\ensuremath{-}0.021}
\newcommand{\VOneTwoDiffci}{\mbox{[\ensuremath{-}0.075, +0.038]}}
\newcommand{\VOneTwoWilcoxon}{0.36}

\newcommand{\SOneSeventeen}{0.428}

\newcommand{\SOneDUSeventeen}{+0.000}

\newcommand{\SOneDSixtyEight}{+0.049}

\newcommand{\SOneDUSixtyEight}{+0.027}

\newcommand{\SOneDOneFifty}{+0.059}

\newcommand{\SOneDUOneFifty}{+0.028}

\newcommand{\SOneOneB}{0.539}

\newcommand{\SOneQwen}{0.398}

\newcommand{\SOneUniform}{0.328}

\newcommand{\SOneDUPooledOneB}{+0.043}

\newcommand{\HolmTaskUSixtyEight}{0.11}

\newcommand{\HolmTaskUOneFifty}{0.16}

\newcommand{\HolmNTaskSig}{6}
\newcommand{\HolmNItemSig}{11}
\newcommand{\SOneVTwoMinusVOne}{\ensuremath{-}0.008}
\newcommand{\SOneVTwoMinusVOneitem}{\mbox{[\ensuremath{-}0.013, \ensuremath{-}0.002]}}

\newcommand{\SOneUnansweredCrivo}{72}

\newcommand{\SOnePrumo}{0.340}
\newcommand{\SOneUPrumo}{0.345}
\newcommand{\BTPrumo}{0.209}
\newcommand{\BTBestPubUnseen}{0.618}

\newcommand{\BTUSixtyEight}{0.553}

\newcommand{\BTUFourHundred}{0.636}

\newcommand{\BTUOneB}{0.686}

\newcommand{\BTDUSeventeenQwen}{+0.017}
\newcommand{\BTDUSeventeenQwenci}{\mbox{[+0.007, +0.027]}}
\newcommand{\BTNUnseen}{4}

\newcommand{\FlipSeventeen}{23}
\newcommand{\FlipOneB}{12}
\newcommand{\FlipDeberta}{2}
\newcommand{\FlipQwen}{76}
\newcommand{\FlipGemma}{67}
\newcommand{\FlipEttin}{67}

\newcommand{\BoardEceMin}{0.080}
\newcommand{\BoardEceMax}{0.151}
\newcommand{\BoardEcePrumo}{0.047}
\newcommand{\SOneBrierUUniform}{0.619}
\newcommand{\PubAurcSeventeen}{0.29}
\newcommand{\PubAurcOneB}{0.16}
\newcommand{\PubEceSeventeen}{0.076}
\newcommand{\PubEceOneB}{0.158}
\newcommand{\GpuSeventeen}{8.2}
\newcommand{\GpuOneB}{24.3}
\newcommand{\GpuFourHundred}{24.1}
\newcommand{\GpuDeberta}{36.1}
\newcommand{\GpuDebertaMean}{119.1}
\newcommand{\GpuDebertaSkipped}{40}
\newcommand{\GpuSeventeenMean}{10.9}

\newcommand{\CpuPrumo}{0.2}

\newcommand{\CpuN}{80}

\newcommand{\CpuSeventeen}{13.5}
\newcommand{\CpuSeventeenMean}{80}
\newcommand{\CpuDeberta}{728}
\newcommand{\CpuDebertaMean}{12,103}
\newcommand{\CpuDebertaRatio}{54}
\newcommand{\CpuJointMean}{139}
\newcommand{\CpuPeroptMean}{3,728}
\newcommand{\CpuJointMedian}{21.9}
\newcommand{\CpuPeroptMedian}{41.0}
\newcommand{\CpuPeroptMedianRatio}{1.9}
\newcommand{\CpuPeroptMeanRatio}{27}
\newcommand{\CpuPeroptPNineFiveRatio}{32}
\newcommand{\RelThirtyTwoPubc}{0.305}
\newcommand{\PPubc}{0.245}
\newcommand{\RelThirtyTwoSOne}{0.447}
\newcommand{\RelSixtyEightSOne}{0.466}
\newcommand{\PMeanSOne}{0.455}

\newcommand{\RelThirtyTwoSOneU}{0.430}

\newcommand{\PMeanSOneU}{0.428}

\newcommand{\RelThirtyTwoBT}{0.597}
\newcommand{\RelSixtyEightBT}{0.662}
\newcommand{\PMeanBT}{0.602}

\newcommand{\RelThirtyTwoBTU}{0.506}

\newcommand{\PMeanBTU}{0.497}

\newcommand{\PubcN}{12,833}
\newcommand{\AOneSkill}{+0.058}
\newcommand{\AOneSkillci}{\mbox{[+0.050, +0.065]}}
\newcommand{\AOneSOne}{+0.037}
\newcommand{\AOneSOneci}{\mbox{[+0.002, +0.073]}}
\newcommand{\AOneSOneU}{+0.022}

\newcommand{\AOneBT}{+0.116}
\newcommand{\AOneBTci}{\mbox{[+0.056, +0.177]}}
\newcommand{\AOneBTU}{+0.058}

\newcommand{\ATwoJnoBrierSkill}{+0.002}
\newcommand{\ATwoJnoBrierSkillci}{\mbox{[\ensuremath{-}0.009, +0.012]}}

\newcommand{\ATwoJnoBrierSOneBrier}{\ensuremath{-}0.009}
\newcommand{\ATwoJnoBrierSOneBrierci}{\mbox{[\ensuremath{-}0.016, \ensuremath{-}0.003]}}

\newcommand{\ATwoJnoKDSkill}{\ensuremath{-}0.003}
\newcommand{\ATwoJnoKDSkillci}{\mbox{[\ensuremath{-}0.006, \ensuremath{-}0.000]}}

\newcommand{\ATwoJnoKDBT}{\ensuremath{-}0.044}
\newcommand{\ATwoJnoKDBTci}{\mbox{[\ensuremath{-}0.190, +0.101]}}

\newcommand{\ATwoJrawSkill}{\ensuremath{-}0.005}

\newcommand{\ATwoJrawSOne}{+0.008}

\newcommand{\ATwoJrawBT}{\ensuremath{-}0.075}
\newcommand{\ATwoJrawBTci}{\mbox{[\ensuremath{-}0.242, +0.091]}}

\newcommand{\ATwoJrawFlip}{\ensuremath{-}0.053}
\newcommand{\ATwoJrawFlipci}{\mbox{[\ensuremath{-}0.086, \ensuremath{-}0.021]}}
\newcommand{\ConeSkill}{\ensuremath{-}0.011}
\newcommand{\ConeSkillci}{\mbox{[\ensuremath{-}0.038, +0.015]}}

\newcommand{\CtwoSkill}{+0.031}

\newcommand{\CtwoSOne}{+0.036}

\newcommand{\CtwoBT}{+0.123}

\newcommand{\CthreeBT}{+0.073}

\newcommand{\ASkillJ}{0.180}
\newcommand{\ASkillJsteps}{0.187}
\newcommand{\ASkillP}{0.245}

\newcommand{\SSOneFiftyVOne}{27.74}

\newcommand{\RPrumoTransfer}{0.77}
\newcommand{\NPrumoTransfer}{34}

\newif\ifhaveJcomp \haveJcomptrue

\newif\ifhavePraw \havePrawtrue

\newif\ifhaveGpuLat \haveGpuLattrue

\newif\ifhaveCpuLat \haveCpuLattrue

\newcommand{\JcompSkill}{0.217}
\newcommand{\AOneCompSkill}{+0.028}
\newcommand{\AOneCompSkillci}{\mbox{[+0.015, +0.042]}}

\newcommand{\AOneCompSOne}{+0.031}
\newcommand{\AOneCompSOneci}{\mbox{[+0.010, +0.051]}}

\newcommand{\AOneCompSOneU}{+0.011}
\newcommand{\AOneCompSOneUci}{\mbox{[\ensuremath{-}0.009, +0.031]}}

\newcommand{\AOneCompBT}{+0.070}
\newcommand{\AOneCompBTci}{\mbox{[+0.018, +0.121]}}

\newcommand{\AOneCompBTU}{+0.025}
\newcommand{\AOneCompBTUci}{\mbox{[\ensuremath{-}0.034, +0.084]}}

\newcommand{\JcompSOneBrier}{0.672}

\newcommand{\AOneCompEce}{+0.063}
\newcommand{\AOneCompEceci}{\mbox{[+0.023, +0.102]}}

\newcommand{\GpuThreeJointMedian}{10.4}
\newcommand{\GpuThreeJointMean}{13.7}

\newcommand{\GpuThreePeroptMedian}{10.7}
\newcommand{\GpuThreePeroptMean}{59.0}

\newcommand{\GpuThreeN}{1,000}

\newcommand{\GpuThreeMeanRatio}{4.3}
\newcommand{\GpuThreePNineFiveRatio}{10}

\newcommand{\TokJcomp}{136.7}
\newcommand{\TokPraw}{161.3}
\newcommand{\TokRatio}{1.18}

\newcommand{\JcompMinusJstepsSkill}{+0.029}
\newcommand{\JcompMinusJstepsSkillci}{\mbox{[+0.016, +0.043]}}
\newcommand{\JcompMinusJstepsSOne}{+0.006}

\newcommand{\JcompMinusJstepsBT}{+0.047}

\newcommand{\PerOptRepo}{\url{https://huggingface.co/tardellirs/crivo-32m-per-option-ablation}}
\newcommand{\CodeRepo}{\url{https://github.com/tardellirs/crivo-decision-encoders}}
\newcommand{\thisjournal}{in IEEE Access}
\newcommand{\di}{Decision Index\xspace}
\newcommand{\opt}{\texttt{[OPT]}\xspace}

\newcommand{\PARstart}[2]{#1#2}
\NewDocumentCommand{\Figure}{O{t!} d() O{} m m}{%
  \begin{figure}[#1]\centering\includegraphics[#3]{#4}\caption{#5}\end{figure}}
\renewenvironment{abstract}
  {\begin{center}\begin{minipage}{0.9\textwidth}\small\noindent\textbf{Abstract.}\ }
  {\end{minipage}\end{center}}
\newenvironment{keywords}
  {\begin{center}\begin{minipage}{0.9\textwidth}\small\noindent\textbf{Index Terms.}\ }
  {\end{minipage}\end{center}\vspace{0.6em}}

\title{Joint or Per-Option? Compact Encoders for Closed-Set Decisions from 17 Million to 1 Billion Parameters}
\author{Tardelli Ronan Coelho Stekel}

\hypersetup{pdftitle={Joint or Per-Option? Compact Encoders for Closed-Set Decisions from 17 Million to 1 Billion
Parameters}, pdfauthor={Tardelli Ronan Coelho Stekel}}

\begin{document}

\twocolumn[{%
\begin{center}
{\LARGE\bfseries Joint or Per-Option? Compact Encoders for Closed-Set Decisions\\[2pt]from 17 Million to 1 Billion
Parameters\par}
\vspace{1em}
{\large Tardelli Ronan Coelho Stekel\par}
\vspace{0.3em}
{\small Federal Institute of S\~ao Paulo (IFSP), S\~ao Paulo, Brazil\\
\texttt{stekel@ifsp.edu.br} \quad ORCID: 0000-0002-3135-3656\par}
\end{center}
\vspace{0.4em}
\begin{abstract}
Many language-model pipelines end in a closed-set decision, such as selecting a tool. Prompting a large language
model for it is slow and order-sensitive. We ask how small an encoder for such decisions can be and whether it should
read the options jointly or separately. In a three-seed study at 32 million parameters, per-option scoring beats joint
scoring on a held-out sample of a 37-benchmark decision suite by 5.8 points of chance-corrected skill at matched
optimizer steps and by 2.8 at approximately matched training compute; two external benchmarks agree, though not
significantly on their unseen tasks. Crossing reader with initialisation shows that the advantage requires a reranker
initialisation, whose query--candidate pair input matches per-option reading; from the raw encoder the two readers are
indistinguishable. Per-option scoring needs 5 to 13 times more encoder passes: at most twice the median latency, but
more than four times the mean. Of six released joint encoders (17 million to 1 billion parameters), the 68- and
150-million-parameter models exceed a 435-million-parameter DeBERTa-v3 zero-shot classifier by 4.9 and 5.9 accuracy
points on 209 overlap-audited System1-Arena tasks and are at least on par with it on the 145 tasks unseen in training.
The smallest answers a median request in 14 milliseconds on 16 processor threads, against 728 for the classifier.
Reordering options changes 12 to 23\% of the joint models' answers (per-option scoring: none). Up to 68 million
parameters, they lead their size range on a public leaderboard's private, mainly in-distribution tests.
\end{abstract}


}]

\section{Introduction}
\label{sec:intro}

\PARstart{M}{any} uses of a language model end in a decision from a closed set: which tool to call, whether a
retrieved passage supports a claim, which intent a message expresses, which of four answers is right. In established
terms these are zero-shot classification and multiple-choice problems whose label set changes from instance to
instance. Current practice answers them in one of three ways. A large language model (LLM) can be prompted and its
output parsed, which is slow and sensitive to the order in which the options are listed
\cite{zheng2024robust,pezeshkpour2024order}. A natural language inference (NLI) model can score each candidate label
as a hypothesis \cite{yin2019benchmarking,laurer2023building}, which needs one forward pass per option. Or a classifier
can be fine-tuned for each task, which does not transfer to new label sets. A recent request format, introduced with
the Jev decision model \cite{almeida2026jev} and formalised by later work \cite{antunes2026ndm,li2026jeva}, packages
such problems as \emph{typed decisions}: a structured request (a \emph{state} plus typed \emph{questions}) goes in, and
a probability for every allowed option comes out, with no generated text (Fig.~\ref{fig:task}).

Because the output space is fixed, a model for such decisions can in principle be small. This paper asks two
questions: how small an encoder can be, and how it should read the options. The second question has two classical
answers. Per-option scoring, as in BERT's multiple-choice head \cite{devlin2019bert,zellers2018swag} and in NLI
zero-shot classification, encodes the context once per option and normalises the scores across options; joint
scoring encodes the context with all options in one sequence and reads one score per option from marker tokens, as in
UniMC \cite{yang2022unimc} and, for typed decisions, Dinah-0 \cite{duarte2026dinah0}. Joint scoring costs one pass per
question instead of $n$, which is why recent compact decision models use it, but the two have not been compared under
controlled conditions for this kind of decision.

\begin{figure*}[!t]
\centering
\includegraphics[width=\textwidth]{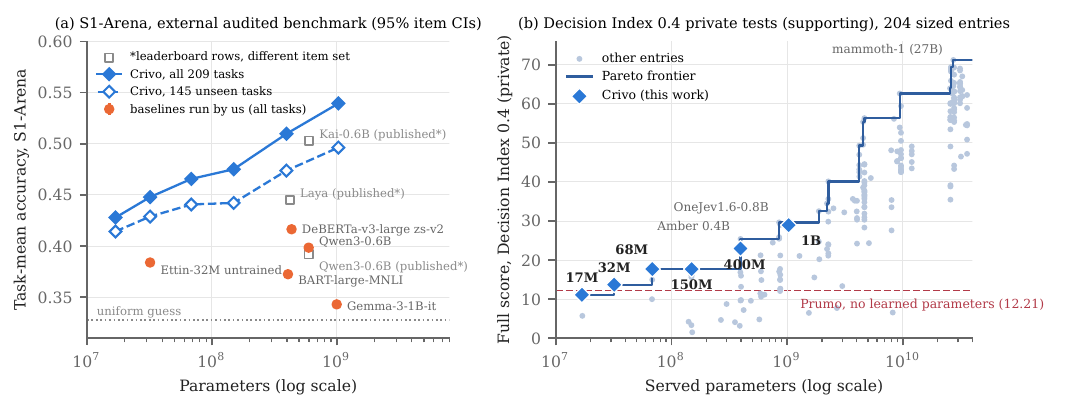}
\caption{\textbf{Compact encoders on closed-set decisions.} (a) S1-Arena, our runs: task-mean accuracy over all
209 tasks (filled) and over the 145 tasks unseen by the Crivo training data (open), with 95\% per-item bootstrap
intervals, against baselines run under the same protocol (orange; the two LLMs use a letter readout without a chat
template) and three published leaderboard rows (squares; full benchmark, not paired). (b) Supporting evidence: \di
0.4, official Full score (private tests only) against served parameters for the \NSized{} sized entries (snapshot of
2026-10-20; board maximum \BoardMaxFull); staircase: size--score Pareto frontier; differences below 0.9 points are
within the board's rank-group rule. Prumo (no learned parameters) is the dashed line.}
\label{fig:overview}
\end{figure*}

We train \textbf{Crivo}, a family of six joint encoders from 17M to 1B parameters fine-tuned from the Ettin rerankers
\cite{weller2025ettin,aarsen2026ettinreranker} with an \opt readout adapted from Dinah-0, and we run a three-seed
controlled grid at 32M that compares joint and per-option reading with everything else fixed, from two
initialisations. The grid is scored on a held-out sample of the 37-benchmark public suite of the \di
\cite{multimodalart2026jevdecisionindex} (hereafter, the board), a public leaderboard for this format
(\emph{public-suite skill}). We evaluate the family on two public external benchmarks audited for overlap with our
training data, System1-Arena (S1-Arena) \cite{chen2026system1arena} for typed decisions and BTZSC \cite{aarab2026btzsc}
for zero-shot text classification, against named baselines run under one protocol, and we measure option-order
robustness, calibration and latency on stated hardware. As supporting evidence we report the board's private tests,
whose official score comes only from tests its maintainers run, together with \textbf{Prumo}, a rule engine with no
learned parameters that shows what those tests reward without learning. Fig.~\ref{fig:overview} summarises the
external and the private-test results.

\gdef\xfigwd{0pt}

\subsection{Contributions}
\begin{enumerate}
\item \textbf{A controlled comparison of joint and per-option reading.} In a 32M grid with three seeds per arm,
per-option scoring beats the joint reader used in the released models, when both start from the reranker, by
\AOneSkill{} public-suite skill at matched optimizer steps and by \AOneCompSkill{} at approximately matched training
compute, with the same sign on S1-Arena and BTZSC but no significant difference on their unseen tasks. A $2\times2$
of initialisation by reader locates the advantage: it appears only from the cross-encoder reranker, whose pair input
matches per-option reading; from the raw encoder the two readers are indistinguishable. Per-option scoring costs
5--13 times more encoder passes, felt mostly in mean and tail latency (Section~\ref{sec:ablation}).
\item \textbf{Six compact joint encoders with contamination-audited external evaluation.} The released models (17M--1B)
took about US\$80 of rented GPU time (US\$54 recorded, the rest estimated). On all 209 S1-Arena tasks, Crivo-68M and
Crivo-150M exceed DeBERTa-v3-large zero-shot (435M) by \SOneDSixtyEight{} and \SOneDOneFifty{} in task-mean accuracy,
also after a Holm correction; on the 145 unseen tasks they are at least on par with it. On the four unseen BTZSC
datasets, Crivo-400M and Crivo-1B match or exceed published zero-shot rows of up to 8B parameters
(Sections~\ref{sec:res-s1} and~\ref{sec:res-btzsc}).
\item \textbf{Robustness, calibration and cost on stated hardware.} Reordering options changes
\FlipOneB--\FlipSeventeen\% of Crivo's answers (DeBERTa \FlipDeberta\%, per-option scoring none); the models are
over-confident on unseen tasks; Crivo-17M answers a median request in \CpuSeventeen~ms on 16 CPU threads, against
\CpuDeberta~ms for DeBERTa (Sections~\ref{sec:res-flip}--\ref{sec:res-eff}).
\item \textbf{Private-test evidence and a probe of the benchmark.} On the board's maintainer-run private tests,
Crivo-17M, -32M and -68M lead their size range on mostly in-distribution skills but stay near the floor on new
domains, and our own development panels, drawn from the same pool as the training data, overestimated every model's
private score (Section~\ref{sec:sel-neg}). Prumo, a rule engine with no learned parameters, matches Crivo-17M on the
same-skill tests while staying near chance externally, which exposes format-specific solving in part of the board
(Section~\ref{sec:selection}).
\end{enumerate}

The task format \cite{almeida2026jev} and the \opt readout \cite{duarte2026dinah0} are prior work. All weights, code,
run outputs and the scripts behind every table and figure are released (Section~\ref{sec:availability}).
Section~\ref{sec:related} reviews related work, Section~\ref{sec:protocol} gives the protocol and
Section~\ref{sec:method} the models. Section~\ref{sec:ablation} reports the reader study, Section~\ref{sec:results}
the external results and Section~\ref{sec:selection} the private tests; Section~\ref{sec:discussion} discusses
implications and limitations.

\section{Related Work}
\label{sec:related}

\subsection{Option scoring: per-option, joint and listwise}
\label{sec:rel-options}
How an encoder should read a set of candidate answers is an old question, and the reader study of
Section~\ref{sec:ablation} belongs to it. \emph{Per-option} scoring encodes the context once per option and normalises
the scores across options. It is the standard multiple-choice head of BERT-style encoders on SWAG and RACE
\cite{devlin2019bert,zellers2018swag,lai2017race}, the basis of zero-shot classification by entailment, which scores
each label as an NLI hypothesis \cite{yin2019benchmarking} using models trained on corpora such as MNLI
\cite{williams2018mnli} (BART-large-MNLI \cite{lewis2020bart} and the DeBERTa-v3 \cite{he2023debertav3} classifiers of
Laurer et~al.\ \cite{laurer2024less,laurer2023building} are the standard checkpoints), and of label-pair classifiers
such as TARS \cite{halder2020tars}. Its cost grows with the number of options, and options cannot be compared
directly, which multiple-choice reading-comprehension models address with option-comparison and co-matching layers
\cite{ran2019ocn,zhang2020dcmn}.

\emph{Joint} scoring puts the context and all options in one sequence and reads one
score per option. UniMC \cite{yang2022unimc}, extended \thisjournal{} \cite{wang2023zero}, recasts natural language
understanding tasks as multiple choice: an option-mask token precedes every option, each predicts ``yes'' through the
masked-language-model head, and the ``yes'' logits are normalised across options; an attention mask keeps the options
from attending to each other, so all options are read in one pass but scored without direct comparison. GLiNER
\cite{zaratiana2024gliner} prepends one marker per entity type and reads all types in one pass; GLiClass
\cite{stepanov2025gliclass} applies the same idea to zero-shot sequence classification, encoding a text with all its
candidate labels, which attend to each other and to the text, and reports several times the throughput of pairwise NLI
cross-encoders; and LACO \cite{zhang2021laco} jointly encodes a document with all labels of a fixed label set for
supervised multi-label classification. Recasting classification as
multiple choice over category labels has also been studied \thisjournal{} \cite{zhang2025comprehensive}.

In retrieval, listwise cross-encoders such as Set-Encoder \cite{schlatt2025setencoder} let candidates attend to each other
and make the result invariant to their order. For LLMs, multiple-choice answers are usually read from the logit of the
option letter \cite{robinson2023leveraging}, which is sensitive to option order \cite{zheng2024robust,
pezeshkpour2024order}. BTZSC compares cross-encoders, embedding models, rerankers and LLMs under one zero-shot protocol
\cite{aarab2026btzsc}.

Crivo's released reader is a joint reader in the UniMC line, adapted to typed decisions through Dinah-0. Unlike
UniMC, and like GLiClass, it lets the options attend to each other; it also uses a fresh head on a reranker backbone
and is trained on typed decisions rather than on converted multiple-choice datasets. Our reader study compares it with per-option
scoring under controlled conditions (Section~\ref{sec:ablation}).

\subsection{Typed decision models}
Jev introduced the request format used here, under the name ``System One'', together with a hosted decision model
\cite{almeida2026jev}. Independent evaluations and an early evidence review followed within weeks
\cite{deusser2026evaluating,duan2026general,sharma2026typedbench,tang2026typed}, among them S1-Arena, a
typed-decision benchmark built from 268 public sources with a public leaderboard \cite{chen2026system1arena}, and a
harness in which a fine-tuned DeBERTa-v3-large wins on almost every labelled benchmark \cite{rafe2026benchmarking}.
Most open decision models are decoders read through option letters or deliberation
\cite{geuter2026releasing,li2026llmasjev,zhang2026anyjev,zhuang2026startlux}. Encoder approaches are closer to ours.
Decision models have been built from biomedical sentence encoders \cite{deka2026retrieval}; Bekko released a large training set and 17M--400M models on the same Ettin rerankers that we start from
\cite{tateno2026bekkodataset,tateno2026bekkomodels}; Laya, a 421M ModernBERT-based model, was independently reproduced
and found under-confident \cite{nandakishore2026laya}; Verdict answers all questions of a request in one pass under a
structured attention mask \cite{chouhan2026verdict}; and Narrow Decision Models and Jeva formalise the interface with a
bidirectional encoder and structural requirements such as permutation invariance \cite{antunes2026ndm,li2026jeva}.
The \opt readout that Crivo uses comes from the Dinah-0 model
\cite{duarte2026dinah0}, a 150M encoder for this format whose option markers resemble UniMC's.

\subsection{Cross-encoders, compact encoders and distillation}
Crivo is a cross-encoder in the sense of monoBERT \cite{nogueira2019passage}, as opposed to the siamese bi-encoders of
Sentence-BERT \cite{reimers2019sentencebert}: the context and the candidates are encoded jointly. Its backbones are the
reranker fine-tunes \cite{aarsen2026ettinreranker} of the Ettin encoders \cite{weller2025ettin}, which follow
ModernBERT \cite{warner2025modernbert}, a successor of BERT \cite{devlin2019bert}. Smaller family members are distilled
from larger ones, as compact encoders often are: with soft targets \cite{hinton2015distilling} as in DistilBERT
\cite{sanh2019distilbert}, by attention distillation as in MiniLM \cite{wang2020minilm}, and in recent task-aware
distillation into encoders \cite{yoo2026token}.

\subsection{Calibration and selective prediction}
We report expected calibration error \cite{naeini2015obtaining,guo2017calibration}, the Brier score
\cite{brier1950verification} and selective-prediction risk \cite{geifman2017selective}. Pre-trained transformers are
reasonably calibrated in domain and less so out of domain \cite{desai2020calibration}; decision models have been found
over-confident when answers need specialist knowledge \cite{duan2026general} and under-confident in the case of Laya
\cite{nandakishore2026laya}.

\subsection{Leaderboards, holdout reuse and contamination}
Reusing a test set for selection overfits it \cite{dwork2015reusable}, models lose accuracy on fresh test sets built to
replicate the original distribution \cite{recht2019imagenet}, and leaderboards can be distorted when some providers
test many private variants and publish only the best \cite{singh2025leaderboard}; contamination audits
\cite{sainz2023nlp} are one response. Our negative result (Section~\ref{sec:sel-neg}) is a case where the proxy that
failed was our own development panel, not the public test set. Following Schwartz et~al.\
\cite{schwartz2020greenai}, we report quality against cost.

\subsection{Rule-based and parameter-free baselines}
Parameter-free classifiers such as compression with nearest neighbours \cite{jiang2023lowresource} and deterministic
tiers that hand over to a model only when needed \cite{lim2026when} show that simple baselines can be strong and
diagnostic. Prumo follows this line: a fixed pipeline of classical components (BM25 \cite{robertson2009probabilistic},
code execution, game-tree search, truth tables, calendar arithmetic) used as a diagnostic of what a decision benchmark
rewards without learning.

\section{Task and Evaluation Protocol}
\label{sec:protocol}

\subsection{Typed decisions}
A request has a \emph{state} (text or JSON: a conversation, a document, a tool catalogue, a board position) and one or
more \emph{questions}, each with instructions. A \emph{multiple-choice} question lists option keys, optionally with
descriptions, and expects a probability for every key. A \emph{binary predicate} question expects the probability that
a statement is true. An \emph{ordinal} question expects a distribution over ordered levels. The answer is the argmax;
probabilities also serve ranking benchmarks and selective prediction. (In the Jev format these types are called
\texttt{choice}, \texttt{noul} and \texttt{score}.) Fig.~\ref{fig:task} shows an invented support ticket with one
question of each type and the answers of a released Crivo model. In a pipeline the encoder sits between the
application or LLM agent that builds the request and a rule that acts on the probabilities: route the ticket, pick a
tool, or block an action when a guardrail question passes a threshold.

\Figure[t!](topskip=0pt, botskip=0pt, midskip=0pt)[width=0.99\columnwidth]{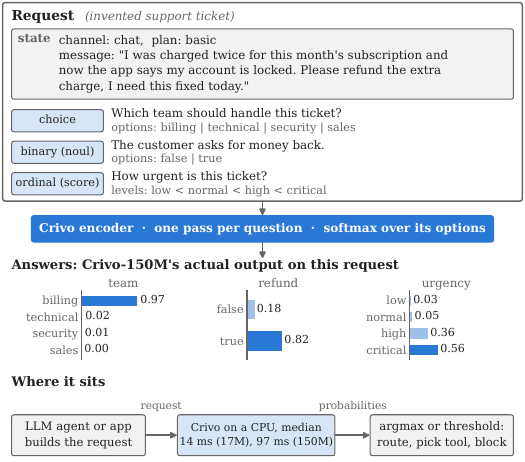}
{A typed decision request and where a compact encoder sits. The invented request holds a state and three typed
questions; each question is one encoder pass and receives a probability for every option. Bars are the output of the
released Crivo-150M on this request, run once on a CPU (an illustration, not a benchmark sample). Latencies are the
16-thread CPU medians of Table~\ref{tab:latency}.\label{fig:task}}

\subsection{External benchmarks}
\label{sec:proto-ext}
\textbf{System1-Arena} (S1-Arena) \cite{chen2026system1arena} is a typed-decision benchmark in the same request
format, with open data and code and a public leaderboard. We use the 13 non-agentic domains of its public snapshot, which its overall leaderboard
score averages: 41,183 requests, 72,817 questions and 209 tasks (the published leaderboard also includes 2,043 requests that can only
be built locally, so its rows are not on exactly our item set). We map binary questions to binary predicates; nothing
else is changed. The statistic is the benchmark's own: per-task accuracy averaged over tasks, computed with its scorer
(unanswered counts as wrong). Every Crivo leaves \SOneUnansweredCrivo{} questions unanswered: they belong to 24
requests longer than the 32,768 tokens that the released engine reads, which it declines rather than truncates.

\textbf{BTZSC} \cite{aarab2026btzsc} is a zero-shot text classification benchmark of 22 datasets in sentiment, emotion,
intent and topic, with published results for NLI cross-encoders, rerankers, embedding models and LLMs. We render each
item as one multiple-choice question whose options are BTZSC's own label verbalisations, with one fixed instruction
chosen before any run. The metric is macro-F1 averaged over datasets. Running Crivo-17M through the official harness
on 50 items per dataset gave per-dataset scores identical to our adapter's.

\subsection{Contamination policy and audit}
\label{sec:contam}
Our policy has three parts.
(i) Training may use the train and development splits of a benchmark's source datasets, never test items. The \di
permits this, our board entries disclose it, and the board counts any test item found in training data as wrong.
(ii) A firewall drops training rows that match a public-suite test row by normalised state and question, by a
distinctive state, or by a shared passage of at least 120 characters; a final audit of the exact training files against
all 563,686 suite questions found no match.
(iii) The external benchmarks were run on 2026-10-20, after every model had been released, and were never used for
training, selection or tuning: one run per model with one fixed rendering. After training, we audited them against every string in the training data
(Table~\ref{tab:contam}). A task or dataset is \emph{unseen} if no training source is the same dataset or corpus and
fewer than 0.5\% of its test items overlap a training string (exact match of at least five words, or at least 80\% of
the item's word 8-grams inside one training string); \emph{seen: split} if its source corpus was in training but the
overlap is below 0.5\%; and \emph{seen: overlap} otherwise. The S1-Arena audit uses each item's longest state string and
therefore flags shared premises as overlap. We report unseen and seen results separately and call only unseen results
evidence of generalisation; nothing in seen sets is called zero-shot.

\begin{table}[!t]
\centering
\caption{Contamination audit of the external benchmarks against the training data of the released Crivo models.}
\label{tab:contam}
\setlength{\tabcolsep}{3pt}
\begin{tabular}{@{}lrrrr@{}}
\toprule
Benchmark (unit) & Unseen & Seen: split & Seen: overlap & Unseen items \\
\midrule
S1-Arena (tasks) & 145 & 25 & 39 & 28,723 / 41,183 \\
BTZSC (datasets) & 4 & 12 & 6 & 18,852 / 172,501 \\
\bottomrule
\end{tabular}

\end{table}

Only \BTNUnseen{} of 22 BTZSC datasets are unseen (appreviews, empathetic, trueteacher, capsotu), so BTZSC is a
secondary validation; 145 of 209 S1-Arena tasks are unseen. The MoritzLaurer \texttt{zeroshot-v2.0} checkpoints were
fine-tuned on training data of all 22 BTZSC datasets, whose label verbalisations BTZSC inherits from the same dataset
pool \cite{laurer2024zeroshotv2,aarab2026btzsc}, so on BTZSC they are a supervised upper reference, not a zero-shot baseline; we did not
audit their exposure to S1-Arena.

\subsection{Baselines}
\label{sec:proto-base}
All baselines are public checkpoints run by us through the same request files:
\begin{itemize}
\item NLI zero-shot: DeBERTa-v3-large-zeroshot-v2.0 (435M) and ModernBERT-large-zeroshot-v2.0 (395M, BTZSC only)
of Laurer et~al.\ \cite{laurer2023building,laurer2024zeroshotv2}, called ``zs-v2'' below, and BART-large-MNLI \cite{lewis2020bart} (407M,
S1-Arena only); premise = rendered state, hypothesis =
question plus option, softmax of the entailment logits over options. As in BTZSC's harness, the premise is truncated at
512 tokens (1,024 for BART).
\item Small LLMs with a letter readout: Qwen3-0.6B \cite{yang2025qwen3} and Gemma-3-1B-it \cite{gemmateam2025gemma3};
next-token probabilities of the option letters, renormalised, with BTZSC's prompt generalised to a question. Both run
without a chat template or permutation debiasing \cite{zheng2024robust}; they measure a plain letter readout, not the
best these models can do (Gemma-3-1B-it is close to the uniform guess on S1-Arena).
\item The untrained Ettin-32M reranker, i.e.\ Crivo-32M before training, scoring each option as a document.
\item A uniform guess.
\end{itemize}
Inputs a baseline cannot process (too long or too many options) count as wrong, as in the \di. We also cite the
published S1-Arena and BTZSC rows, which come from the benchmarks' own harnesses and carry no paired intervals. We
did not run the compact decision encoders of other authors
(Bekko, Dinah-1, the tasksource decider, Mola) or GLiClass \cite{stepanov2025gliclass} on the external benchmarks, and
none of them publishes results on S1-Arena or BTZSC that we could compare with; they appear only on the \di.

\subsection{The Decision Index public suite}
\label{sec:proto-di}
The \di \cite{multimodalart2026jevdecisionindex,apolinario2026decisionindex} is a public leaderboard for this format.
Entrants run a public suite of 37 benchmarks in five areas (knowledge and reasoning, language understanding, retrieval
and classification, tools and automation, arts and taste) with the open kit and submit their outputs. Each benchmark
is chance-corrected, $(s-c)/(1-c)$ clipped to $[0,1]$, and the \emph{public index} is 100 times the weighted mean. The
suite covers tool use \cite{patil2025bfcl,ross2025when2call,li2023apibank,shi2025toolret}, intent classification
\cite{casanueva2020banking77,larson2019clinc}, fact verification \cite{jiang2020hover,niu2024ragtruth}, NLI
\cite{nie2020anli,koreeda2021contractnli,jullien2024nli4ct}, sentiment and stance
\cite{cai2021acos,tang2023finentity,allaway2020vast,abufarha2022isarcasmeval}, commonsense and knowledge
\cite{zellers2019hellaswag,sakaguchi2020winogrande,wang2024mmlupro,rein2023gpqa,phan2025hle}, reasoning and code
\cite{suzgun2023bbh,cobbe2021gsm8k,gu2024cruxeval,jin2023cladder,sprague2024musr,ruoss2024chessbench}, retrieval
\cite{reddy2022esci,su2024bright}, and arts and taste \cite{hossain2019humicroedit,hessel2023androids,yao2025bachi,
tessler2024habermas}. The board's official ranking uses only private tests run by its maintainers
(Section~\ref{sec:selection}).

\subsection{Statistics}
\label{sec:proto-stats}
For S1-Arena and BTZSC we use a per-item paired bootstrap: requests are resampled with replacement inside each task or
dataset (the questions of a request move together), the official statistic is recomputed, and the same draws serve all
systems ($B=2{,}000$). This interval is conditional on the benchmark's tasks. A second, wider interval resamples tasks
within domains (S1-Arena) or datasets (BTZSC) and speaks to ``tasks like these''. For the twelve comparisons of the six Crivo models
with DeBERTa (all and unseen tasks) we also give Holm-adjusted bootstrap $p$-values. The ablation grid uses Welch 95\%
intervals over three seeds per arm; with 3 against 3 seeds they have 2--4 degrees of freedom and assume normality,
and an exact permutation test could not go below $p=0.10$, so we treat intervals that barely exclude zero as
suggestive. Correlations across board entries use a bootstrap over entries ($B=10{,}000$); entries are not independent
(several are fine-tunes of one base), and we did not cluster the bootstrap. Tables and text give accuracy, skill and their
differences as fractions; where we speak of ``points'' (abstract, introduction) a point is 0.01 of such a fraction,
while board scores are on a 0--100 scale.

\subsection{Robustness and latency measurement}
\label{sec:proto-rob}
Option-order robustness uses 3,841 multiple-choice questions (2,833 requests) sampled from the public suite: each is
answered under three random permutations and under reversed order, and 258 binary requests are also answered with
their two options swapped. We report the share of questions whose chosen option changes; requests a system cannot
process (121 for DeBERTa, 200 for the two LLMs) are excluded. Latency is measured one request at a time on the first
1,000 requests of a seeded shuffle of this sample, the same for every system: on one unshared NVIDIA A100-SXM4 40\,GB
(bf16 autocast, 20 warm-up requests), and on 16 threads of an AMD EPYC 7J13 with nothing else running (fp32, 10
warm-up requests, first \CpuN{} requests, all answered by every system). Each request is timed once, except in a
paired GPU run of the per-option checkpoint P (Section~\ref{sec:ablation}) and Crivo-32M, whose two alternating passes are averaged. Crivo batches up
to 32 questions of a request per call, the NLI baseline up to 32 premise--hypothesis pairs, and the per-option reader
all options of a request in shared batches. On the GPU, requests a system cannot process are not timed
(\GpuDebertaSkipped{} of 1,000 for DeBERTa, which flatters it).

\section{Models}
\label{sec:method}

\subsection{Reader}
Crivo follows the \opt readout of Dinah-0 \cite{duarte2026dinah0}, which in turn resembles the option-mask tokens of
UniMC \cite{yang2022unimc}: one marker token per option inside a single sequence, scored and normalised across the
options. Each question becomes one sequence,
\begin{center}
\texttt{[CLS] instr [SEP] state [SEP] [OPT] $o_1$ \dots\ [OPT] $o_n$},
\end{center}
where an option $o_i$ is rendered as \texttt{key: description}, or as \texttt{key} alone. A binary predicate becomes
two options (false, true) and an ordinal question one option per level; JSON is rendered compactly with sorted keys.
\opt is a new token initialised from \texttt{[SEP]}. The hidden state $h_i$ at the $i$-th \opt, plus a learned
embedding $e_t$ of the question type $t$, goes through a small head (LayerNorm, Linear, GELU, Linear; in fp32) to one
logit, and
\begin{equation}
p(o_i \mid x) = \frac{\exp f(h_i + e_t)}{\sum_{j=1}^{n} \exp f(h_j + e_t)} .
\end{equation}
Options attend to each other, so a question costs one encoder pass whatever its number of options; we call this the
\emph{joint} reader.

The \emph{per-option} reader of the reader study (Section~\ref{sec:ablation}) encodes $n$ sequences
\texttt{[CLS] instr [SEP] state [SEP] [OPT] $o_i$}, with the same head and the same softmax across them
(Fig.~\ref{fig:reader}); it is the multiple-choice head of BERT \cite{devlin2019bert} with an \opt marker in place of
\texttt{[CLS]}.

The state is tokenised once per request and all questions of a request are batched. The backbones use rotary positions with alternating local (128-token) and global attention; models are trained up to 8,192
tokens and read inputs whole up to 32,768 tokens, so nothing is truncated at inference. Option order is shuffled in
training; the released engine reads one order.

\Figure[t!](topskip=0pt, botskip=0pt, midskip=0pt)[width=0.99\columnwidth]{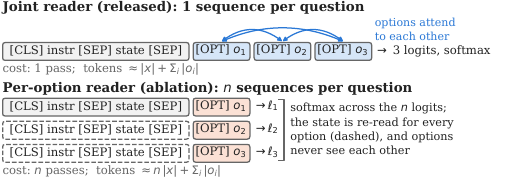}
{The two readers compared in Section~\ref{sec:ablation}. Both apply the same head at the \opt positions and a
softmax over the question's options. The joint reader (released) encodes one sequence in which the options attend
to each other; the per-option reader encodes $n$ sequences, re-reading the instruction and state ($|x|$ tokens) once
per option, and its options never see each other. Token counts per question are approximate (markers omitted).\label{fig:reader}}

\subsection{Model family}
All models start from the Ettin reranker of the same size (\texttt{cross-encoder/ettin-reranker-*-v1})
\cite{aarsen2026ettinreranker}; Table~\ref{tab:models} lists them. The six released models share the recipe below.
Two models trained with an earlier recipe, Crivo-150M v1 and Crivo-32M v1, appear only as the teacher of Crivo-17M
(Crivo-150M v1) and in the recipe comparison of Section~\ref{sec:sel-neg}; unmarked names refer to the released
models.

\begin{table}[!t]
\centering
\caption{The Crivo models. Parameters are the board's served count; teachers distil stages 1--2 only. Cost: rented
GPU time for training plus evaluation.}
\label{tab:models}
\setlength{\tabcolsep}{2.2pt}
\begin{tabular}{@{}lrrll@{}}
\toprule
Model & Params & Layers/dim & Teacher & Cost (US\$) \\
\midrule
Crivo-17M & 16.9M & 7 / 256 & Crivo-150M v1 & 13--16 (with 68M)$^a$ \\
Crivo-32M & 32.0M & 10 / 384 & Crivo-400M & 10 (with 150M)$^a$ \\
Crivo-68M & 68.4M & 19 / 512 & Crivo-400M & (with 17M) \\
Crivo-150M & 149.6M & 22 / 768 & Crivo-400M & (with 32M) \\
Crivo-400M & 395.8M & 28 / 1024 & none & 28 \\
Crivo-1B & 1.03B & 28 / 1792 & none & 26 \\
\midrule
Crivo-32M v1 & 32.0M & 10 / 384 & Crivo-150M v1 & 7 \\
Crivo-150M v1 & 149.6M & 22 / 768 & none & 14 \\
\bottomrule
\end{tabular}

\smallskip
\raggedright\footnotesize $^a$Estimated; each pair was trained in one shared A100 session ($\sim$7\,h and $\sim$5\,h).
Other costs are recorded cloud bills.
\end{table}

\subsection{Training data}
The training mix has 638,306 questions in stage~1, 169,437 in stage~2 and 24,635 in stage~3. It combines the Bekko System
One data \cite{tateno2026bekkodataset} (at most 1,200 rows per subset), 328 tasksource families
\cite{sileo2024tasksource,tasksource2026jevtyped} (40 further families held out entirely), procedural typed decisions
\cite{tasksource2026procedural,sileo2026reasoningcore}, code-generated decision families with exact labels (policies
with exceptions, opaque label rules, schedules, ordinal severity, dates, propositional logic, executed code), synthetic
tool-use dialogues built from public function-calling data, and the train splits of the public suite's source
datasets, capped at 5,000 rows per source, about half rewritten into other request layouts. No outputs of Jev or of other proprietary
decision APIs were used as targets. The API-Bank \cite{li2023apibank} train split repeats dialogues of the board's test
pool; it is excluded from the released recipe but was in the training data of Crivo-150M v1, the teacher of
Crivo-17M.

\subsection{Objective, stages and selection}
Each question is a softmax over its options. The loss is cross-entropy plus $0.5\times$ the Brier score; with a teacher
it is $\alpha(\text{CE}+0.5\,\text{Brier})+(1-\alpha)\tau^2\,\mathrm{KL}(p^{\text{teacher}}_\tau\|p^{\text{student}}_\tau)$
with $\alpha=0.5$ and $\tau=2$ for more than two options ($\tau=1$ otherwise); some sources train on gold labels only.
Stage~1 trains on the main mix at up to 2,048 tokens with a weight EMA \cite{tarvainen2017meanteachers}; stage~2
continues on a smaller mix; stage~3 trains on long inputs (up to 8,192 tokens) with gold labels only, so long rows are
moved rather than cut. We use AdamW (weight decay 0.01), linear warm-up (5\% in stage~1, 3\% later) and decay, bf16
and separate learning rates for encoder and head, the head's 5--10 times the encoder's; the models take 5.9k--16.3k
optimizer steps in stage~1, 2.3k--5.2k in stage~2 and 2.6k--5.2k in stage~3, with 16k--33k tokens per step and
encoder learning rates from $10^{-4}$ (17M and 32M, stage~1) down to $5\times10^{-6}$ (1B, stage~3). The per-stage
values are in the model cards.

Checkpoints (end of stage~2 or~3, their average, or an interpolation toward the base model
\cite{wortsman2022wiseft}) were chosen on internal panels: development sets for 8 of the board's skills built from
sibling sources, 21 held-out decision families, and a sample of the public suite. The complete public suite was run
once, on the final checkpoint, and each model was submitted to the board once, after its checkpoint was fixed. The
board influenced the recipe once: the released recipe was designed in response to the gap between the public and
private scores of the v1 models (Section~\ref{sec:sel-neg}), a mild form of the adaptive holdout reuse that Dwork
et~al.\ \cite{dwork2015reusable} warn about. Every model was trained on one rented GPU (A100 40\,GB or H100 80\,GB);
costs are in Table~\ref{tab:models}.

\subsection{Prumo: a floor with no learned parameters}
Prumo reads the same requests and returns the same typed answers, but contains no model, no embeddings and no corpus
statistics. Each question passes through twelve capabilities in a fixed order: sandboxed code execution, chess search
(alpha--beta, 7,000 nodes), BM25 tool relevance with schema validation, BM25 retrieval, lexical grounding, a consensus
centroid, IDF label overlap, chord templates, exact Boolean logic and SAT, exhaustive puzzle search (claimed only if the
solution is unique), calendar arithmetic, and exact-fraction arithmetic. A capability either claims the question and
returns probabilities or passes it on; unclaimed questions get the uniform distribution, so their expected score is
exactly chance. The engine never reads benchmark names or item ids. It is about 2.7k lines of Python with one
dependency (\texttt{python-chess}); its only numbers are 527 hand-set constants, 454 of them chess piece-square
values. Thresholds were chosen on train and development splits. All budgets are counted in operations, so outputs are
bit-identical across machines. Prumo executes code taken from requests in a sandbox with resource limits and should not
be exposed to untrusted input without further isolation.

\section{Reader Study: Joint versus Per-Option Scoring}
\label{sec:ablation}

\subsection{Design}
\label{sec:abl-design}
Joint reading lets options attend to each other in one pass, which could help comparison; per-option reading re-reads
the context for every option but matches the (query, document) format on which cross-encoder rerankers are trained.
To compare them with everything else fixed, we trained a grid at 32M on a fixed subset of 150,000 questions sampled
from the stage-1 mix, for one epoch, with the same rows, hyperparameters (those of Crivo-32M stage~1), Crivo-400M
teacher logits and evaluation sets for every arm, and three seeds per arm. The arms are: \textbf{J}, the released
reader and objective (reranker initialisation, CE~+~0.5\,Brier~+~distillation); \textbf{P}, the per-option reader
(Fig.~\ref{fig:reader}) with everything else equal; \textbf{J without Brier}; \textbf{J without distillation}; and
\textbf{J-raw}, J from the raw Ettin encoder instead of the reranker. Evaluation uses sets that are never trained on: a
stratified 15\% sample of the public suite (\emph{pub15}; 13,358 requests; chance-corrected skill averaged over the 37
benchmarks, \emph{not} the official index), 60 items per S1-Arena task (\emph{S1-60}; 12,540 requests), BTZSC's
500-item harness sample (\emph{BTZSC-500}; 11,000 requests), and the order perturbations of
Section~\ref{sec:proto-rob}. The arms and metrics were fixed before the grid ran.

Because each per-option sequence holds a single option, P takes about 3.25 times more optimizer steps than J at the
same per-step token budget (6,858 against 2,112 on the same rows). Two joint controls separate the reader from this
difference. \textbf{J-steps} shrinks the per-step budget so that J takes P's number of steps; it matches updates, not
compute, and was added after the grid had started. \textbf{J-comp}, the primary comparison, keeps J's flags and
per-step budget and trains for exactly P's number of steps in each seed (6,858--6,860) with the same schedule length.
It reads each row about 3.25 times, so it is a repetition control rather than a fresh-data control, and it is close
to but not exactly matched in encoded tokens (\TokJcomp M against \TokPraw M, \TokRatio{} times) and GPU time
(per-option runs took about four times longer because batches were padded to the longest sequence). The primary
endpoint is the Welch interval of P $-$ J-comp on pub15 skill. A $2\times2$ of initialisation (reranker or raw Ettin
encoder) by reader adds \textbf{P-raw} and \textbf{J-comp-raw}, three seeds each, with contrasts C1 (P-raw $-$
J-comp-raw), C2 (P $-$ P-raw), C3 (J-comp $-$ J-comp-raw) and their interaction C4 $=$ C2 $-$ C3
(Welch--Satterthwaite over the four cells). The plans and decision rules of J-comp and the $2\times2$ were recorded in
a dated experiment log before the runs (released with the code). The $2\times2$ cells come from three runs on the same
GPU type with identical code, data, flags and scorers, so a run effect cannot be separated. P lost 2, 0 and 6 of about
6,860 training batches to out-of-memory errors; no other arm lost any.

\begin{table*}[!t]
\centering
\caption{Reader study at 32M (150k questions, one epoch except J-comp and J-comp-raw, three seeds per arm; arms defined
in Section~\ref{sec:abl-design}). Top: mean $\pm$ standard deviation over seeds. Lower blocks: differences with Welch
95\% intervals over seeds (bold: interval excludes 0). ``--'': flips not measured (per-option reading has none by
construction). $^\ddagger$Not a grid arm: the released model's outputs on the same requests (pub15 on the \PubcN{}
requests in its public-suite run; P scores \PPubc{} there). Lower is better for Brier, ECE and flips.}
\label{tab:ablation}
\scriptsize
\setlength{\tabcolsep}{2.4pt}
\begin{tabular}{@{}lrrrrrrrr@{}}
\toprule
Arm & pub15 skill & S1-60 acc. & S1-60 unseen & BTZSC-500 F1 & unseen-4 F1 & S1-60 Brier & pub15 ECE & Perm. flips \\
\midrule
\multicolumn{9}{@{}l}{\textit{Mean $\pm$ s.d. over 3 seeds}} \\
J (released reader) & 0.180$\pm$0.002 & 0.406$\pm$0.003 & 0.401$\pm$0.003 & 0.425$\pm$0.069 & 0.386$\pm$0.062 & 0.612$\pm$0.002 & 0.123$\pm$0.017 & 0.277$\pm$0.015 \\
J-steps (steps matched to P) & 0.187$\pm$0.004 & 0.417$\pm$0.014 & 0.406$\pm$0.009 & 0.485$\pm$0.029 & 0.439$\pm$0.020 & 0.602$\pm$0.002 & 0.102$\pm$0.036 & 0.262$\pm$0.030 \\
J-comp (compute matched to P) & 0.217$\pm$0.006 & 0.424$\pm$0.008 & 0.417$\pm$0.009 & 0.532$\pm$0.006 & 0.472$\pm$0.011 & 0.672$\pm$0.008 & 0.040$\pm$0.014 & 0.253$\pm$0.009 \\
P (per-option) & 0.245$\pm$0.003 & 0.455$\pm$0.001 & 0.428$\pm$0.003 & 0.602$\pm$0.022 & 0.497$\pm$0.028 & 0.584$\pm$0.003 & 0.103$\pm$0.019 & 0.000$\pm$0.000 \\
J-comp-raw (J-comp from raw encoder) & 0.226$\pm$0.013 & 0.416$\pm$0.001 & 0.406$\pm$0.002 & 0.459$\pm$0.011 & 0.431$\pm$0.033 & 0.676$\pm$0.004 & 0.056$\pm$0.039 & -- \\
P-raw (P from raw encoder) & 0.214$\pm$0.011 & 0.419$\pm$0.004 & 0.408$\pm$0.002 & 0.478$\pm$0.014 & 0.403$\pm$0.006 & 0.608$\pm$0.008 & 0.108$\pm$0.008 & 0.000 \\
J without Brier & 0.182$\pm$0.005 & 0.425$\pm$0.011 & 0.420$\pm$0.007 & 0.483$\pm$0.049 & 0.448$\pm$0.028 & 0.603$\pm$0.003 & 0.135$\pm$0.021 & 0.265$\pm$0.034 \\
J without distillation & 0.177$\pm$0.001 & 0.405$\pm$0.004 & 0.398$\pm$0.007 & 0.380$\pm$0.031 & 0.363$\pm$0.034 & 0.608$\pm$0.004 & 0.127$\pm$0.009 & 0.276$\pm$0.002 \\
J-raw (J from raw encoder) & 0.175$\pm$0.004 & 0.413$\pm$0.001 & 0.405$\pm$0.007 & 0.350$\pm$0.010 & 0.311$\pm$0.029 & 0.608$\pm$0.003 & 0.149$\pm$0.019 & 0.223$\pm$0.007 \\
Crivo-32M released (joint, full recipe)$^\ddagger$ & 0.305 & 0.447 & 0.430 & 0.597 & 0.506 & 0.609 & -- & 0.196 \\
\midrule
\multicolumn{9}{@{}l}{\textit{Differences, Welch 95\% CI over seeds (3 vs 3)}} \\
P $-$ J-comp (matched compute) & \shortstack{\textbf{+0.028}\\{\tiny\mbox{[+0.015, +0.042]}}} & \shortstack{\textbf{+0.031}\\{\tiny\mbox{[+0.010, +0.051]}}} & \shortstack{+0.011\\{\tiny\mbox{[\ensuremath{-}0.009, +0.031]}}} & \shortstack{\textbf{+0.070}\\{\tiny\mbox{[+0.018, +0.121]}}} & \shortstack{+0.025\\{\tiny\mbox{[\ensuremath{-}0.034, +0.084]}}} & \shortstack{\textbf{\ensuremath{-}0.088}\\{\tiny\mbox{[\ensuremath{-}0.106, \ensuremath{-}0.070]}}} & \shortstack{\textbf{+0.063}\\{\tiny\mbox{[+0.023, +0.102]}}} & \shortstack{\textbf{\ensuremath{-}0.253}\\{\tiny\mbox{[\ensuremath{-}0.275, \ensuremath{-}0.231]}}} \\
P $-$ J-steps (matched steps) & \shortstack{\textbf{+0.058}\\{\tiny\mbox{[+0.050, +0.065]}}} & \shortstack{\textbf{+0.037}\\{\tiny\mbox{[+0.002, +0.073]}}} & \shortstack{\textbf{+0.022}\\{\tiny\mbox{[+0.002, +0.042]}}} & \shortstack{\textbf{+0.116}\\{\tiny\mbox{[+0.056, +0.177]}}} & \shortstack{\textbf{+0.058}\\{\tiny\mbox{[+0.002, +0.115]}}} & \shortstack{\textbf{\ensuremath{-}0.019}\\{\tiny\mbox{[\ensuremath{-}0.024, \ensuremath{-}0.013]}}} & \shortstack{+0.001\\{\tiny\mbox{[\ensuremath{-}0.072, +0.075]}}} & \shortstack{\textbf{\ensuremath{-}0.262}\\{\tiny\mbox{[\ensuremath{-}0.336, \ensuremath{-}0.188]}}} \\
\midrule
\multicolumn{9}{@{}l}{\textit{Initialisation $2\times2$, Welch 95\% CI over seeds}} \\
C1: P-raw $-$ J-comp-raw & \shortstack{\ensuremath{-}0.011\\{\tiny\mbox{[\ensuremath{-}0.038, +0.015]}}} & \shortstack{+0.003\\{\tiny\mbox{[\ensuremath{-}0.007, +0.013]}}} & \shortstack{+0.002\\{\tiny\mbox{[\ensuremath{-}0.003, +0.007]}}} & \shortstack{+0.020\\{\tiny\mbox{[\ensuremath{-}0.010, +0.050]}}} & \shortstack{\ensuremath{-}0.028\\{\tiny\mbox{[\ensuremath{-}0.107, +0.050]}}} & \shortstack{\textbf{\ensuremath{-}0.068}\\{\tiny\mbox{[\ensuremath{-}0.084, \ensuremath{-}0.052]}}} & \shortstack{+0.053\\{\tiny\mbox{[\ensuremath{-}0.039, +0.144]}}} & -- \\
C2: P $-$ P-raw & \shortstack{\textbf{+0.031}\\{\tiny\mbox{[+0.006, +0.055]}}} & \shortstack{\textbf{+0.036}\\{\tiny\mbox{[+0.026, +0.046]}}} & \shortstack{\textbf{+0.020}\\{\tiny\mbox{[+0.014, +0.026]}}} & \shortstack{\textbf{+0.123}\\{\tiny\mbox{[+0.077, +0.169]}}} & \shortstack{\textbf{+0.094}\\{\tiny\mbox{[+0.030, +0.159]}}} & \shortstack{\textbf{\ensuremath{-}0.025}\\{\tiny\mbox{[\ensuremath{-}0.041, \ensuremath{-}0.008]}}} & \shortstack{\ensuremath{-}0.005\\{\tiny\mbox{[\ensuremath{-}0.046, +0.036]}}} & -- \\
C3: J-comp $-$ J-comp-raw & \shortstack{\ensuremath{-}0.009\\{\tiny\mbox{[\ensuremath{-}0.035, +0.017]}}} & \shortstack{+0.007\\{\tiny\mbox{[\ensuremath{-}0.013, +0.028]}}} & \shortstack{+0.011\\{\tiny\mbox{[\ensuremath{-}0.010, +0.032]}}} & \shortstack{\textbf{+0.073}\\{\tiny\mbox{[+0.051, +0.095]}}} & \shortstack{+0.041\\{\tiny\mbox{[\ensuremath{-}0.031, +0.113]}}} & \shortstack{\ensuremath{-}0.005\\{\tiny\mbox{[\ensuremath{-}0.021, +0.012]}}} & \shortstack{\ensuremath{-}0.015\\{\tiny\mbox{[\ensuremath{-}0.100, +0.070]}}} & -- \\
C4: C2 $-$ C3 (interaction) & \shortstack{\textbf{+0.040}\\{\tiny\mbox{[+0.014, +0.066]}}} & \shortstack{\textbf{+0.028}\\{\tiny\mbox{[+0.011, +0.045]}}} & \shortstack{+0.009\\{\tiny\mbox{[\ensuremath{-}0.009, +0.028]}}} & \shortstack{\textbf{+0.050}\\{\tiny\mbox{[+0.006, +0.094]}}} & \shortstack{+0.053\\{\tiny\mbox{[\ensuremath{-}0.015, +0.121]}}} & \shortstack{\textbf{\ensuremath{-}0.020}\\{\tiny\mbox{[\ensuremath{-}0.038, \ensuremath{-}0.002]}}} & \shortstack{+0.010\\{\tiny\mbox{[\ensuremath{-}0.066, +0.086]}}} & -- \\
\bottomrule
\end{tabular}

\end{table*}

\Figure[t!](topskip=0pt, botskip=0pt, midskip=0pt)[width=0.99\columnwidth]{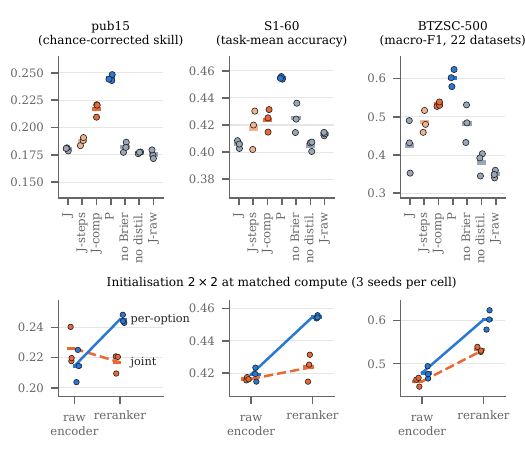}
{Reader study at 32M: per-arm mean (horizontal tick) and the three seeds (dots) on pub15, S1-60 and BTZSC-500. Top: P
(per-option, blue) is above every joint arm on all three, including J-comp (orange), which matches P's optimizer steps
and per-step token budget, and J-steps (light orange), which matches only its steps. Bottom: the initialisation
$2\times2$ at matched compute, per-option (P-raw to P, blue, solid) and joint (J-comp-raw to J-comp, orange, dashed):
the per-option reader gains from the reranker on all three sets, the joint reader only on BTZSC-500. Axes do not start
at zero.\label{fig:ablation}}

\subsection{Per-option against joint reading}
Table~\ref{tab:ablation} and Fig.~\ref{fig:ablation} give the result. At matched steps, the per-option reader is better
on every evaluation set: pub15 skill \AOneSkill{} \AOneSkillci, S1-60 task-mean accuracy \AOneSOne{} \AOneSOneci{}
(\AOneSOneU{} on unseen tasks) and BTZSC-500 macro-F1 \AOneBT{} \AOneBTci{} (\AOneBTU{} on the unseen datasets). The
S1-60 and BTZSC-500 intervals are wide or barely exclude zero and should be read as consistent with the pub15 gain
rather than as separate confirmations. Matching steps closes little of the gap: J-steps improves on J only from
\ASkillJ{} to \ASkillJsteps{} pub15 skill, while P reaches \ASkillP.

\textbf{Compute-matched control.} The gap is not only a compute effect. J-comp, which trains the joint reader for as
many steps and about as many tokens as P, reaches \JcompSkill{} pub15 skill, and P remains ahead (\AOneCompSkill{}
\AOneCompSkillci; S1-60 \AOneCompSOne{} \AOneCompSOneci; BTZSC-500 \AOneCompBT{} \AOneCompBTci). Compute closes about
half of the step-matched pub15 gap (J-comp $-$ J-steps \JcompMinusJstepsSkill{} \JcompMinusJstepsSkillci) but less on
S1-60 (\JcompMinusJstepsSOne) and BTZSC-500 (\JcompMinusJstepsBT), where the intervals include zero. The advantage is
not significant on unseen data: on the unseen S1-Arena tasks and the unseen BTZSC datasets the compute-matched
differences are \AOneCompSOneU{} \AOneCompSOneUci{} and \AOneCompBTU{} \AOneCompBTUci. Calibration does not follow
accuracy. J-comp has the lowest pub15 ECE of all arms (P $-$ J-comp \AOneCompEce{} \AOneCompEceci) but a worse S1-60
Brier score (\JcompSOneBrier, against 0.602 for J-steps), as does J-comp-raw; both read each row about 3.25 times,
which fits over-confidence from repetition.

\textbf{Per benchmark.} The compute-matched pub15 gain is concentrated (Fig.~\ref{fig:a1bench}). P is better on 22
benchmarks and worse on 10; the paired interval excludes zero for 8 gains and one loss (ContractNLI, $-0.127$); the
median gain is only $+0.009$. Five benchmarks (BPoMP, iSarcasmEval, API-Bank, MuSR and ACOS, $+0.12$ to $+0.33$)
carry most of the mean ($+0.015$ without the top two), and on two of them (ACOS and iSarcasmEval) J-comp falls below J,
again consistent with repetition. The gain does not grow with the number of options or with input length (Spearman
$\rho=-0.14$ and $-0.06$ across benchmarks), and the joint arms' position bias (accuracy 0.015 lower with the gold
option first or last) is too small to explain it. On S1-60 P is better on 125 of 209 tasks, on BTZSC-500 on 19 of 22
datasets.

\Figure[t!](topskip=0pt, botskip=0pt, midskip=0pt)[width=0.99\columnwidth]{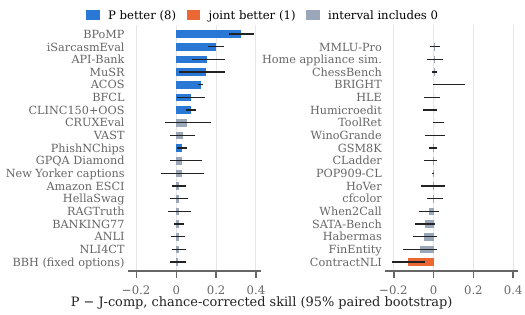}
{Per-benchmark difference P $-$ J-comp (compute-matched) in chance-corrected skill on the 37 benchmarks of pub15 (seed
means of three seeds per arm; bars sorted, continued in the right panel; lines: 95\% paired cluster-bootstrap
intervals, requests resampled within benchmark). A few benchmarks carry most of the mean gain; the joint reader is
better on ContractNLI.\label{fig:a1bench}}

\textbf{Initialisation.} The $2\times2$ (Table~\ref{tab:ablation}, bottom; Fig.~\ref{fig:ablation}, bottom) shows
that the per-option advantage depends on the starting checkpoint. From the raw Ettin encoder the two readers are
indistinguishable at matched compute (C1 contains zero on all three sets; pub15 \ConeSkill{} \ConeSkillci). The
reranker helps the per-option reader on all three sets (C2: \CtwoSkill, \CtwoSOne, \CtwoBT) but the joint reader only
on BTZSC-500 (C3: \CthreeBT), and the interaction C4 is positive on all three. The most direct reading is a format
match: Ettin rerankers are trained on (query, document) pairs, exactly the per-option input, whereas a sequence holding
all options is new to the checkpoint under either initialisation. The $2\times2$ does not test this mechanism, and with
three seeds per cell it cannot exclude a small advantage of either reader at raw initialisation; it does show that at
32M and matched compute there is no evidence that reading options separately is better in itself. On pub15 the path
from J to P thus decomposes into a small step effect (\ASkillJ{} to \ASkillJsteps), a compute effect
(\JcompMinusJstepsSkill) and a per-option advantage of similar size (\AOneCompSkill) that exists only with the
reranker initialisation.

\textbf{Against the released model.} The released Crivo-32M (joint) was trained for two epochs on the full 638k-row
stage-1 mix and then a second stage, about five times more distinct rows than P's one epoch on 150k rows. On the
grid's subsets (Table~\ref{tab:ablation}, $\ddagger$) the released model is clearly ahead on pub15 (\RelThirtyTwoPubc{}
against \PPubc), whose source datasets fill much of its mix, but P matches it on the external ones: S1-60
\PMeanSOne{} against \RelThirtyTwoSOne{} (unseen \PMeanSOneU{} against \RelThirtyTwoSOneU), BTZSC-500 \PMeanBT{}
against \RelThirtyTwoBT{} (unseen-4 \PMeanBTU{} against \RelThirtyTwoBTU); Crivo-68M scores \RelSixtyEightSOne{} and
\RelSixtyEightBT. These unpaired comparisons across very different budgets suggest, but do not show, that a
per-option model trained from the reranker with the full recipe would be the more accurate 32M model; we leave that
run to future work (Section~\ref{sec:conclusion}).

\subsection{Cost}
\label{sec:abl-cost}
Per request, the per-option reader needs 5.2 times more forward sequences on pub15, 6.4 times on S1-60 and 13.1 times
on BTZSC-500, because it re-encodes the state once per option. We timed the per-option checkpoint P (seed 0) against
the released joint Crivo-32M, which has the same architecture, on the same requests (Table~\ref{tab:latency}, paired
block). On the A100, over \GpuThreeN{} requests in two alternating passes, the medians are the same
(\GpuThreePeroptMedian{} against \GpuThreeJointMedian~ms), but the mean is \GpuThreeMeanRatio{} times higher
(\GpuThreePeroptMean{} against \GpuThreeJointMean~ms) and the 95th percentile \GpuThreePNineFiveRatio{} times. On 16
CPU threads the median request takes \CpuPeroptMedian{} against \CpuJointMedian~ms (\CpuPeroptMedianRatio{} times),
the mean \CpuPeroptMean{} against \CpuJointMean~ms (\CpuPeroptMeanRatio{} times) and the 95th percentile
\CpuPeroptPNineFiveRatio{} times longer. Typical requests are cheap either way; the cost falls on long requests with
many options. Per-option training with the full recipe would cost about 3.8--4.2 times the joint reader's tokens on
the main stages and 23 times on the long-input stage (catalogues of about 44 tools share one state). The choice is
therefore a trade-off. Where many options or questions share one long state (tool catalogues, intent lists with
77--151 labels), the joint reader is far cheaper. Where a reranker checkpoint is available and mean and tail latency
are acceptable, the per-option reader is more accurate, even at approximately matched training compute. From a raw
encoder, per-option reading bought no accuracy in our runs.

\subsection{Objective and initialisation of the joint reader}
\label{sec:abl-a2}
The remaining factors have small effects at this scale (arm means in Table~\ref{tab:ablation}; differences and
intervals here). Removing the Brier term changes pub15 skill by \ATwoJnoBrierSkill{} \ATwoJnoBrierSkillci{} and lowers
the S1-60 Brier score slightly (\ATwoJnoBrierSOneBrier{} \ATwoJnoBrierSOneBrierci), so we cannot claim that the Brier
term improves calibration. Removing distillation costs \ATwoJnoKDSkill{} \ATwoJnoKDSkillci{} skill; its BTZSC-500
change (\ATwoJnoKDBT{} \ATwoJnoKDBTci) is within the seed noise. J-raw, the joint reader from the raw Ettin encoder at
J's budget, leaves pub15 skill (\ATwoJrawSkill) and S1-60 accuracy (\ATwoJrawSOne) almost unchanged, lowers order
sensitivity (\ATwoJrawFlip{} \ATwoJrawFlipci) and gives the lowest BTZSC-500 score of all arms (\ATwoJrawBT{}
\ATwoJrawBTci).

\section{External Results}
\label{sec:results}

\subsection{S1-Arena: external typed decisions}
\label{sec:res-s1}
Table~\ref{tab:s1} gives task-mean accuracy on all, unseen and seen tasks, and the difference to the strongest
baseline, DeBERTa-v3-large zs-v2, with both intervals.

\begin{table*}[!t]
\centering
\caption{S1-Arena, task-mean accuracy (the benchmark's statistic) on its public snapshot, with 95\% per-item paired
bootstrap intervals. Unseen/seen follow the audit of Section~\ref{sec:contam}. The last two columns give the difference
to DeBERTa-v3-large zs-v2 with the wider task interval; $^\ast$ the per-item interval excludes 0 (its
half-width is about 0.006), $^\dagger$ the task interval excludes 0. Intervals are not adjusted for multiplicity (see
text).}
\label{tab:s1}
\scriptsize
\setlength{\tabcolsep}{3pt}
\begin{tabular}{@{}lrllrll@{}}
\toprule
 & & \multicolumn{3}{c}{Task-mean accuracy} & \multicolumn{2}{c}{$\Delta$ vs DeBERTa zs-v2 [task CI]} \\
\cmidrule(lr){3-5}\cmidrule(l){6-7}
System & Params & All (209) [item CI] & Unseen (145) [item CI] & Seen (64) & All tasks & Unseen tasks \\
\midrule
\textbf{Crivo-17M} & 16.9M & 0.428 \mbox{[0.423, 0.432]} & 0.414 \mbox{[0.409, 0.420]} & 0.459 & +0.011$^\ast$ \mbox{[\ensuremath{-}0.013, +0.037]} & +0.000 \mbox{[\ensuremath{-}0.021, +0.021]} \\
\textbf{Crivo-32M} & 32.0M & 0.448 \mbox{[0.444, 0.452]} & 0.429 \mbox{[0.424, 0.434]} & 0.491 & +0.031$^\ast$$^\dagger$ \mbox{[+0.007, +0.057]} & +0.015$^\ast$ \mbox{[\ensuremath{-}0.005, +0.035]} \\
\textbf{Crivo-68M} & 68.4M & 0.466 \mbox{[0.461, 0.470]} & 0.441 \mbox{[0.435, 0.446]} & 0.522 & +0.049$^\ast$$^\dagger$ \mbox{[+0.024, +0.074]} & +0.027$^\ast$$^\dagger$ \mbox{[+0.004, +0.049]} \\
\textbf{Crivo-150M} & 149.6M & 0.475 \mbox{[0.471, 0.480]} & 0.442 \mbox{[0.437, 0.447]} & 0.550 & +0.059$^\ast$$^\dagger$ \mbox{[+0.031, +0.086]} & +0.028$^\ast$$^\dagger$ \mbox{[+0.001, +0.055]} \\
\textbf{Crivo-400M} & 395.8M & 0.510 \mbox{[0.505, 0.514]} & 0.474 \mbox{[0.468, 0.479]} & 0.591 & +0.093$^\ast$$^\dagger$ \mbox{[+0.066, +0.120]} & +0.060$^\ast$$^\dagger$ \mbox{[+0.032, +0.086]} \\
\textbf{Crivo-1B} & 1.03B & 0.539 \mbox{[0.535, 0.544]} & 0.496 \mbox{[0.491, 0.502]} & 0.637 & +0.123$^\ast$$^\dagger$ \mbox{[+0.095, +0.151]} & +0.082$^\ast$$^\dagger$ \mbox{[+0.055, +0.109]} \\
\textbf{Crivo-150M v1} & 149.6M & 0.483 \mbox{[0.478, 0.487]} & 0.448 \mbox{[0.442, 0.453]} & 0.561 & +0.066$^\ast$$^\dagger$ \mbox{[+0.039, +0.093]} & +0.034$^\ast$$^\dagger$ \mbox{[+0.007, +0.060]} \\
\midrule
DeBERTa-v3-large zs-v2 & 435.0M & 0.416 \mbox{[0.412, 0.420]} & 0.414 \mbox{[0.409, 0.419]} & 0.422 & -- & -- \\
BART-large-MNLI & 407.0M & 0.373 \mbox{[0.369, 0.376]} & 0.375 \mbox{[0.370, 0.380]} & 0.367 & -- & -- \\
Qwen3-0.6B (letter) & 596.0M & 0.398 \mbox{[0.394, 0.403]} & 0.402 \mbox{[0.397, 0.407]} & 0.390 & -- & -- \\
Gemma-3-1B-it (letter) & 1.00B & 0.343 \mbox{[0.339, 0.347]} & 0.345 \mbox{[0.340, 0.350]} & 0.339 & -- & -- \\
Ettin-32M reranker (untrained) & 32.0M & 0.384 \mbox{[0.380, 0.388]} & 0.388 \mbox{[0.383, 0.393]} & 0.376 & -- & -- \\
Uniform & -- & 0.328 \mbox{[0.324, 0.332]} & 0.334 \mbox{[0.329, 0.339]} & 0.313 & -- & -- \\
\bottomrule
\end{tabular}

\end{table*}

\textbf{All tasks.} On all 209 tasks every Crivo beats every baseline with the item interval excluding zero, and
every pair except Crivo-17M against DeBERTa also with the task interval. After a Holm adjustment over the twelve
Crivo--DeBERTa comparisons (all and unseen tasks), \HolmNTaskSig{} remain significant at the task level (68M--1B on
all tasks, 400M and 1B on unseen tasks) and \HolmNItemSig{} at the item level. Accuracy rises with size, least between
68M and 150M (+0.009, item interval only), and reaches \SOneOneB{} at 1B.

\textbf{Unseen tasks.} On the 145 unseen tasks the margins over DeBERTa shrink to about 0.03 for the mid-size
models and are fragile. Crivo-68M (\SOneDUSixtyEight) and Crivo-150M (\SOneDUOneFifty) are ahead with both intervals
excluding zero, but their task-level lower bounds are near zero and neither survives the Holm adjustment
($p=\HolmTaskUSixtyEight$ and \HolmTaskUOneFifty). Crivo-32M is ahead at the item level only, and Crivo-17M ties
(\SOneDUSeventeen) with 25 times fewer parameters. The result also depends on aggregation: pooled over questions
rather than averaged over tasks, a few large unseen tasks on which DeBERTa is strong dominate, Crivo-17M and Crivo-68M
fall below it, Crivo-150M ties, and only Crivo-400M and Crivo-1B stay ahead (\SOneDUPooledOneB{} for 1B). For models
up to 150M we therefore read the unseen-task result as parity with a 435M zero-shot classifier, not as a gain. The
seen--unseen gap also widens with size (0.459 against 0.414 at 17M, 0.637 against 0.496 at 1B), so part of the larger
models' advantage comes from tasks whose source corpora were in training.

\textbf{Published rows.} The strongest sub-1B published rows, Decision-2.0-Kai-0.6B (0.503), Laya 421M (0.445) and
Qwen3-0.6B (0.392), are on the full benchmark and cannot be paired with our runs (our own Qwen3-0.6B run gives
\SOneQwen). Prumo reaches \SOnePrumo{} (unseen \SOneUPrumo), close to
the uniform guess (\SOneUniform): its exact solvers rarely fire on this benchmark's formats.

\subsection{BTZSC: zero-shot text classification}
\label{sec:res-btzsc}
Table~\ref{tab:btzsc} separates the four datasets unseen in training from the 18 whose source corpora (and,
for six of them, a small share of test items) were seen.

\begin{table}[!t]
\centering
\caption{BTZSC, macro-F1 averaged over datasets. Unseen-4 = appreviews, empathetic, trueteacher, capsotu; seen-18 =
the rest (seen by the Crivo training data, not zero-shot for Crivo). Intervals: 95\% per-item bootstrap. Published
rows come from the benchmark's own runs and are not paired with ours. $^\S$Trained on the train splits of all 22
datasets with BTZSC's label verbalisations: a supervised reference, not zero-shot.}
\label{tab:btzsc}
\scriptsize
\setlength{\tabcolsep}{2.5pt}
\begin{tabular}{@{}lrlll@{}}
\toprule
System & Params & Unseen-4 [item CI] & Seen-18 & All-22 \\
\midrule
\multicolumn{5}{@{}l}{\textit{This work (seen-18 = training split seen)}} \\
\textbf{Crivo-17M} & 16.9M & 0.469 \mbox{[0.460, 0.476]} & 0.523 & 0.513 \\
\textbf{Crivo-32M} & 32.0M & 0.514 \mbox{[0.506, 0.521]} & 0.622 & 0.603 \\
\textbf{Crivo-68M} & 68.4M & 0.553 \mbox{[0.545, 0.561]} & 0.693 & 0.667 \\
\textbf{Crivo-150M} & 149.6M & 0.564 \mbox{[0.554, 0.572]} & 0.695 & 0.671 \\
\textbf{Crivo-400M} & 395.8M & 0.636 \mbox{[0.627, 0.643]} & 0.713 & 0.699 \\
\textbf{Crivo-1B} & 1.03B & 0.686 \mbox{[0.677, 0.693]} & 0.765 & 0.751 \\
\midrule
\multicolumn{5}{@{}l}{\textit{Zero-shot baselines run by us (same adapter)}} \\
Qwen3-0.6B (letter) & 596.0M & 0.451 \mbox{[0.443, 0.459]} & 0.319 & 0.343 \\
Gemma-3-1B-it (letter) & 1.00B & 0.265 \mbox{[0.261, 0.270]} & 0.394 & 0.370 \\
Ettin-32M reranker (untrained) & 32.0M & 0.358 \mbox{[0.350, 0.364]} & 0.349 & 0.351 \\
\midrule
\multicolumn{5}{@{}l}{\textit{Published BTZSC zero-shot baselines (no CI)}} \\
e5-mistral-7b-instruct & 7B & 0.618 & 0.566 & 0.576 \\
Qwen3-Reranker-8B & 8B & 0.608 & 0.748 & 0.722 \\
Qwen3-Embedding-8B & 8B & 0.589 & 0.592 & 0.591 \\
gte-reranker-modernbert-base & 149M & 0.577 & 0.579 & 0.578 \\
e5-base-v2 & 110M & 0.566 & 0.603 & 0.597 \\
Qwen3-Reranker-0.6B & 600M & 0.540 & 0.620 & 0.606 \\
bart-large-mnli & 407M & 0.539 & 0.501 & 0.508 \\
Qwen3-4B & 4B & 0.531 & 0.675 & 0.649 \\
deberta-v3-large-nli & 434M & 0.481 & 0.616 & 0.591 \\
ms-marco-MiniLM-L6-v2 & 22M & 0.395 & 0.428 & 0.422 \\
gemma-3-1b-it & 1B & 0.347 & 0.362 & 0.359 \\
\midrule
\multicolumn{5}{@{}l}{\textit{Supervised upper reference (trained on all 22 datasets' train splits)}} \\
DeBERTa-v3-large zs-v2$^\S$ & 435.0M & 0.744 \mbox{[0.734, 0.752]} & 0.813 & 0.800 \\
ModernBERT-large zs-v2$^\S$ & 395.0M & 0.717 \mbox{[0.709, 0.724]} & 0.789 & 0.776 \\
\bottomrule
\end{tabular}

\end{table}

On the unseen datasets, Crivo-400M (\BTUFourHundred) and Crivo-1B (\BTUOneB) match or exceed the published BTZSC
zero-shot rows, the best being an instruction-tuned 7B embedding model (\BTBestPubUnseen) and an 8B reranker (0.608);
the margin of Crivo-400M over the best row is small, and the comparison is unpaired across separate runs. Crivo-68M
(\BTUSixtyEight) is above the published DeBERTa-v3-large NLI cross-encoder (0.481) and BART-large-MNLI (0.539) in the
same unpaired sense. Every Crivo, Crivo-17M included, beats the zero-shot baselines we ran (Qwen3-0.6B, Gemma-3-1B,
untrained Ettin) on unseen-4 with the item interval excluding zero; the tightest case is Crivo-17M against Qwen3-0.6B,
\BTDUSeventeenQwen{} \BTDUSeventeenQwenci. Unseen-4 has only four datasets and their spread is large, so BTZSC is
secondary evidence. The supervised zs-v2 models beat every Crivo (Crivo-1B trails DeBERTa zs-v2 by 0.058 on unseen-4 and by
0.047 on seen-18); this is the gap to a supervised classifier of about Crivo-400M's size trained on in-task data.

\subsection{Robustness to option order}
\label{sec:res-flip}
Fig.~\ref{fig:flips} gives the share of questions whose chosen option changes under the perturbations of
Section~\ref{sec:proto-rob}, against model size. Reversing the order gives rates within 0.02 of the permutations for
every Crivo (0.127 to 0.242), and swapping the two options of a binary request changes 1--8\% of Crivo's answers.

\Figure[t!](topskip=0pt, botskip=0pt, midskip=0pt)[width=0.97\columnwidth]{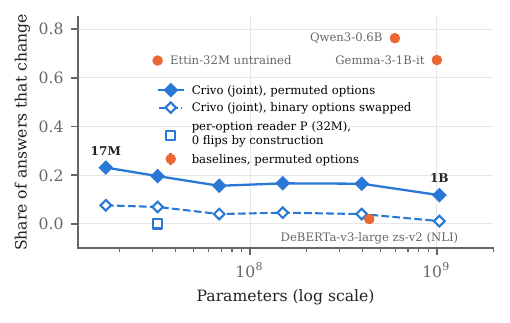}
{Option-order robustness against size: share of answers that change when the options of 3,841 questions are permuted
(mean of three random permutations, 95\% cluster-bootstrap intervals) and, for Crivo, when the two options of 258
binary requests are swapped. The per-option reader is the 32M arm P of Section~\ref{sec:ablation} (three seeds).
Prumo (rules, no size) is invariant.\label{fig:flips}}

Crivo answers change in \FlipOneB--\FlipSeventeen\% of questions, falling with size. Per-option scorers do much
better: the NLI cross-encoder changes only \FlipDeberta\% of its answers, and the per-option reader of
Section~\ref{sec:ablation} none, by construction, because each option is read alone. The two letter-readout LLMs
change \FlipGemma--\FlipQwen\% and the untrained reranker \FlipEttin\%; for the LLMs this is an upper bound, since
permutation debiasing reduces such sensitivity \cite{zheng2024robust}. The NLI cross-encoder is not exactly invariant;
we did not trace its residual changes. Crivo's joint reader sees the order, which is the price of reading all options
in one pass.

\subsection{Calibration}
\label{sec:res-calib}
Table~\ref{tab:calib} collects the board's calibration measurement (maintainer-computed on a 1-in-6 request sample)
and ours on the external benchmarks (expected calibration error with 10 equal-mass bins, ECE, averaged with equal
weight per task or dataset; and the multi-class Brier score).

\begin{table}[!t]
\centering
\caption{Calibration. Board ECE: maintainer-computed. S1-Arena and BTZSC: our runs, ECE with 10 equal-mass bins and
multi-class Brier score, each task or dataset weighted equally. Lower is better. A near-constant system (uniform) has
a low ECE without being useful, and binned ECE on small tasks is biased upward.}
\label{tab:calib}
\scriptsize
\setlength{\tabcolsep}{2.4pt}
\begin{tabular}{@{}lrrrrrrr@{}}
\toprule
 & Board & \multicolumn{3}{c}{S1-Arena} & \multicolumn{3}{c}{BTZSC} \\
\cmidrule(lr){3-5}\cmidrule(l){6-8}
 & & \multicolumn{2}{c}{all 209} & unseen & all-22 & \multicolumn{2}{c}{unseen-4} \\
\cmidrule(lr){3-4}\cmidrule(lr){5-5}\cmidrule(lr){6-6}\cmidrule(l){7-8}
System & ECE & ECE & Brier & Brier & ECE & ECE & Brier \\
\midrule
\textbf{Crivo-17M} & 0.080 & 0.069 & 0.603 & 0.621 & 0.029 & 0.061 & 0.616 \\
\textbf{Crivo-32M} & 0.133 & 0.125 & 0.610 & 0.633 & 0.022 & 0.013 & 0.577 \\
\textbf{Crivo-68M} & 0.103 & 0.133 & 0.595 & 0.625 & 0.040 & 0.087 & 0.520 \\
\textbf{Crivo-150M} & 0.115 & 0.174 & 0.617 & 0.658 & 0.086 & 0.161 & 0.560 \\
\textbf{Crivo-400M} & 0.121 & 0.164 & 0.593 & 0.640 & 0.059 & 0.066 & 0.446 \\
\textbf{Crivo-1B} & 0.151 & 0.191 & 0.579 & 0.636 & 0.049 & 0.043 & 0.395 \\
\midrule
DeBERTa zs-v2 & -- & 0.090 & 0.632 & 0.638 & 0.099 & 0.092 & 0.351 \\
Qwen3-0.6B & -- & 0.229 & 0.712 & 0.706 & 0.257 & 0.084 & 0.668 \\
Gemma-3-1B-it & -- & 0.417 & 0.936 & 0.947 & 0.363 & 0.386 & 0.959 \\
Uniform & -- & 0.038 & 0.616 & 0.619 & -- & -- & -- \\
\bottomrule
\end{tabular}

\end{table}

Crivo's calibration depends on whether it has seen the source corpora: its ECE is lowest where most datasets were
seen in training, and it is over-confident on unseen tasks. On all 22 BTZSC datasets, 18 of them seen in training, every Crivo has a
lower ECE than every baseline (item intervals exclude zero); on the four unseen datasets this no longer holds:
Qwen3-0.6B has a lower ECE than Crivo-68M and Crivo-150M, and DeBERTa than Crivo-150M. On S1-Arena, Crivo-17M has a
lower ECE than DeBERTa ($-0.021$ $[-0.027,-0.015]$) and every Crivo a lower ECE than the two LLMs, but ECE does not
fall with size and the larger Crivos are worse than DeBERTa. On unseen S1-Arena tasks every Crivo has a Brier score no
better than the uniform guess (\SOneBrierUUniform): the models are over-confident on new tasks. On the public suite,
ECE rises from \PubEceSeventeen{} at 17M to \PubEceOneB{} at 1B, while the probabilities rank each model's own errors
better as it grows (area under the risk--coverage curve \PubAurcSeventeen{} to \PubAurcOneB). Board ECE for the Crivos
is \BoardEceMin--\BoardEceMax, in the middle of the small-model field.

\subsection{Efficiency}
\label{sec:res-eff}
Table~\ref{tab:latency} reports latency per request under the protocol of Section~\ref{sec:proto-rob}.

\begin{table}[!t]
\centering
\caption{Latency per request (ms), one request at a time, on fixed public-suite requests (Section~\ref{sec:proto-rob}).
A100: 1,000 requests. CPU: AMD EPYC 7J13, 16 threads, the same first \CpuN{} requests for every system; a few very long
requests dominate the CPU means and 95th percentiles. Paired block: the per-option checkpoint P (seed 0) of
Section~\ref{sec:ablation} and Crivo-32M timed together on the GPU (the ratios of Section~\ref{sec:abl-cost}); their
CPU cells repeat the single CPU run. On the GPU, DeBERTa could not process \GpuDebertaSkipped{} of the 1,000 requests
(not timed). Board: the maintainers' measurement (RTX PRO 6000).}
\label{tab:latency}
\scriptsize
\setlength{\tabcolsep}{2.1pt}
\begin{tabular}{@{}lrrrrrrr@{}}
\toprule
 & \multicolumn{3}{c}{A100 GPU, ms} & \multicolumn{3}{c}{CPU, 16 threads, ms} & Board \\
\cmidrule(lr){2-4}\cmidrule(lr){5-7}
System & med. & mean & p95 & med. & mean & p95 & med. (ms) \\
\midrule
Crivo-17M & 8.2 & 10.9 & 24.9 & 13.5 & 80 & 486 & 5.9 \\
Crivo-32M & 10.5 & 13.9 & 33.3 & 21.9 & 139 & 856 & 7.6 \\
Crivo-68M & 17.6 & 22.5 & 52.3 & 53.2 & 384 & 2,173 & 12.4 \\
Crivo-150M & 19.8 & 27.5 & 78.4 & 97.3 & 684 & 3,859 & 13.4 \\
Crivo-400M & 24.1 & 37.5 & 129.3 & 253.4 & 1,434 & 7,554 & 16.7 \\
Crivo-1B & 24.3 & 50.2 & 225.8 & 578.6 & 3,123 & 19,457 & 16.4 \\
DeBERTa zs-v2 & 36.1 & 119.1 & 552.2 & 728.3 & 12,103 & 61,022 & -- \\
Ettin-32M untrained & 9.5 & 148.7 & 890.9 & 38.5 & 3,750 & 28,202 & -- \\
Prumo (rules) & -- & -- & -- & -- & -- & -- & 0.3 \\
\midrule
\multicolumn{8}{@{}l}{\textit{Paired GPU run: same A100, two alternating passes}} \\
Crivo-32M (joint) & 10.4 & 13.7 & 32.9 & 21.9 & 139 & 856 & -- \\
32M per-option (P) & 10.7 & 59.0 & 328.7 & 41.0 & 3,728 & 27,771 & -- \\
\bottomrule
\end{tabular}

\end{table}

Crivo-17M answers in a median \GpuSeventeen~ms on the GPU and \CpuSeventeen~ms on 16 CPU threads (means
\GpuSeventeenMean{} and \CpuSeventeenMean~ms). DeBERTa-v3-large zs-v2, which needs one NLI pass per option, takes a
median \GpuDeberta~ms on the GPU and \CpuDeberta~ms on the CPU (means \GpuDebertaMean{} and \CpuDebertaMean~ms), about
\CpuDebertaRatio{} times Crivo-17M's CPU median on the same requests. At batch size one, GPU latency is dominated by
fixed overhead: Crivo-400M and Crivo-1B have almost the same median (\GpuFourHundred{} and \GpuOneB~ms), so GPU medians say
little about model size and the CPU columns are the better guide to compute. The median hides a heavy tail of long
tool-retrieval requests with up to 32 questions, visible in the means. Prumo answers in a median of \CpuPrumo~ms on one
CPU thread (1,000 requests, measured separately; not comparable with the 16-thread column).

Fig.~\ref{fig:latency} places each system's S1-Arena accuracy against its median CPU latency on the same requests.
Along the Crivo family, accuracy rises by about 0.11 (\SOneSeventeen{} to \SOneOneB) for a 43-fold increase in
median latency, roughly linearly in its logarithm. Every Crivo up to 1B has a lower median latency than DeBERTa-v3-large
zs-v2, and every one is at least as accurate on all 209 tasks; the untrained Ettin-32M reranker, scored per option, has a
median between those of Crivo-32M and Crivo-68M but is well below both in accuracy.

\Figure[t!](topskip=0pt, botskip=0pt, midskip=0pt)[width=0.97\columnwidth]{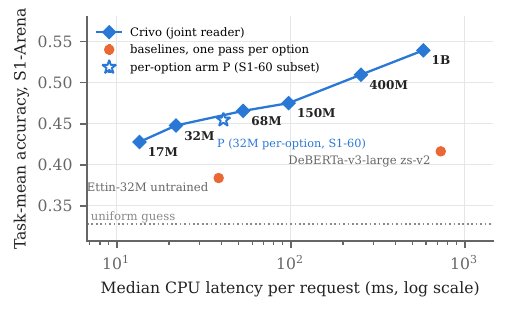}
{Accuracy against CPU cost: S1-Arena task-mean accuracy over all 209 tasks (95\% per-item intervals) against the
median latency per request on 16 CPU threads, on the same \CpuN{} public-suite requests for every system (log
scale); dotted: uniform guess. The per-option arm P has no full S1-Arena run; its hollow star is its accuracy on
S1-60 (mean and range of three seeds), where the released Crivo-32M scores \RelThirtyTwoSOne{} (0.448 on
all items).\label{fig:latency}}

\section{Supporting Evidence: Private Tests of the Decision Index}
\label{sec:selection}

The board's maintainers run each entrant, with the entrant's own serving code, on two private parts: tests ``of
equivalent competence for most public tasks, built from fresh data'' (\emph{same skills}, $S$) and decision tasks from
\emph{new domains} ($O$). In edition 0.4 (October 2026) the ranking score is
\begin{equation}
\text{Full} = 0.60\,S' + 0.40\,O' ,
\end{equation}
where $S'$ and $O'$ are $S$ and $O$ ``mapped onto the public scale with untuned stock models''; the public index is
shown but not counted. The mapping is not documented, but on our snapshot it is affine for all \MapN{} entries (our
fit: \mbox{$S'=\MapSSlope\,S\MapSIntercept$}, \mbox{$O'=\MapOSlope\,O+\MapOIntercept$}, maximum deviations \MapSMaxDev{} and
\MapOMaxDev), so rankings and correlations are unchanged and we report the displayed $S$ and $O$. Entries within 0.9
points of the first model of a group share a \emph{rank group} (\NGroups{} groups); the board publishes no per-item
private data, so rank groups are the only uncertainty we can attach to private scores. Latency is measured by the
maintainers on one NVIDIA RTX PRO 6000 (96~GB), one request at a time. All Full, $S$ and $O$ values are the board's
(edition 0.4, snapshot of 2026-10-20); public indices come from our complete runs of the official kit, which the board
adopted.

Two properties limit what the board can show for our models. First, the board permits training on the train splits
of the suite's source datasets, and Crivo does so (Section~\ref{sec:method}), so $S$, which uses fresh items for the
same tasks, measures in-distribution transfer to fresh items for Crivo, much as BTZSC does for the zs-v2 baselines;
only $O$ measures new domains. Second, the field is sparse at our sizes: \NSmall{} of the \NSized{} sized entries have
at most 420M parameters, \NSmallCrivo{} of them ours, and most of the others are undocumented uploads. We therefore
treat the board as supporting evidence, after the external benchmarks.

\subsection{Official scores}
\label{sec:res-di}
Table~\ref{tab:di} gives the official results of the six released models and Prumo, next to the board entries
closest in size, two stock checkpoints served by their entrants, and the other rule-based entry.

\begin{table*}[!t]
\centering
\caption{Decision Index 0.4, official scores (board snapshot of 2026-10-20). Full $=0.6\,S'+0.4\,O'$ on the board's
mapped private scores; $S$ and $O$ are the displayed private same-skill and new-domain scores; ``Public'' is shown but
not counted. $S$ uses fresh items for tasks whose train splits Crivo trained on. Latency is median / p95 per request on
one RTX PRO 6000, measured by the maintainers. ``Group'': rank group (Section~\ref{sec:selection}). Our entries in
bold.}
\label{tab:di}
\footnotesize
\begin{tabular}{@{}llrrrrcr@{}}
\toprule
Model & Params & \textbf{Full 0.4} & Same skills $S$ & New domains $O$ & Public & Latency (ms) & Group \\
\midrule
\multicolumn{8}{@{}l}{\textit{$\leq$20M}} \\
\textbf{Crivo-17M} & 16.9M & \textbf{11.11} & 15.88 & 3.22 & 22.15 & 5.9 / 10.0 & 47 \\
Bekko v0 17M & 17.0M & 5.72 & 5.89 & 4.61 & 7.44 & 6.8 / 16.5 & 51 \\
\midrule
\multicolumn{8}{@{}l}{\textit{32M}} \\
\textbf{Crivo-32M} & 32.0M & \textbf{13.70} & 18.44 & 5.34 & 27.34 & 7.6 / 11.6 & 45 \\
\midrule
\multicolumn{8}{@{}l}{\textit{68M}} \\
\textbf{Crivo-68M} & 68.4M & \textbf{17.69} & 25.27 & 5.03 & 32.82 & 12.4 / 17.6 & 42 \\
Mola-68M & 68.3M & 14.96 & 21.99 & 3.52 & 27.27 & 11.4 / 18.0 & 44 \\
Bekko v0 68M & 68.0M & 9.99 & 12.62 & 4.98 & 13.01 & 11.8 / 30.6 & 48 \\
\midrule
\multicolumn{8}{@{}l}{\textit{150M}} \\
\textbf{Crivo-150M} & 149.6M & \textbf{17.70} & 24.93 & 5.48 & 34.38 & 13.4 / 24.5 & 42 \\
Mola-150M & 149.3M & 16.15 & 23.55 & 4.03 & 31.34 & 12.8 / 21.1 & 43 \\
Dinah-1 & 149.6M & 16.08 & 22.29 & 5.45 & 34.32 & 14.2 / 57.6 & 43 \\
tasksource nano v4 & 149.0M & 15.63 & 22.08 & 4.79 & 30.90 & 14.2 / 22.6 & 43 \\
\midrule
\multicolumn{8}{@{}l}{\textit{400M}} \\
\textbf{Crivo-400M} & 395.8M & \textbf{22.96} & 30.70 & 9.09 & 39.79 & 16.7 / 39.9 & 39 \\
Amber 0.4B & 395.9M & 25.41 & 35.73 & 7.86 & 49.74 & 42.5 / 47.2 & 37 \\
DecisionTune 1.0 & 394.8M & 16.88 & 22.11 & 7.31 & 28.85 & 13.8 / 111.0 & 42 \\
Bekko v0 400M & 400.0M & 16.00 & 21.55 & 6.20 & 22.45 & 17.1 / 72.2 & 43 \\
Laya & 421.3M & 7.21 & 9.08 & 3.69 & 6.19 & 5.8 / 222.5 & 50 \\
\midrule
\multicolumn{8}{@{}l}{\textit{0.6--1B}} \\
\textbf{Crivo-1B} & 1.03B & \textbf{28.96} & 38.75 & 11.39 & 47.77 & 16.4 / 70.7 & 35 \\
OneJev1.6-0.8B & 853.0M & 29.66 & 40.49 & 10.65 & 48.76 & 13.1 / 103.9 & 35 \\
StartLux-Decision v1.1 0.8B & 873.4M & 29.02 & 41.29 & 8.35 & 53.79 & 29.9 / 87.1 & 35 \\
vLLM-SR Decision 2.0 Kai 0.6B & 600.0M & 14.89 & 16.90 & 9.72 & 16.68 & 18.5 & 44 \\
\midrule
\multicolumn{8}{@{}l}{\textit{stock checkpoints served by their entrants (no own training)}} \\
jeff & 575.7M & 4.81 & 6.86 & 1.53 & 8.27 & 21.8 / 67.6 & 52 \\
Qwen-2.5-1B-RLCD & 1.54B & 6.48 & 6.21 & 5.78 & 4.02 & 43.8 / 75.2 & 51 \\
\midrule
\multicolumn{8}{@{}l}{\textit{no learned parameters}} \\
\textbf{Prumo} & -- & \textbf{12.21} & 15.78 & 5.60 & 20.14 & 0.3 / 11.5 & 46 \\
Zev & -- & 1.91 & 1.79 & 1.89 & 1.55 & 0.5 / 1.7 & 54 \\
\bottomrule
\end{tabular}

\end{table*}

\textbf{Size range.} No entry by another author with fewer than \FirstAboveSixtyEightParams{} parameters outscores
Crivo-68M (\FullSixtyEight); the closest are DecisionTune 1.0 (16.88, in its rank group), Bekko v0 400M (16.00) and
Mola-68M (14.96). Crivo-17M, the smallest sized entry, starts the size--Full Pareto frontier, which continues through
Crivo-32M, Mola-68M and Crivo-68M (Fig.~\ref{fig:overview}b). The frontier positions of Crivo-150M and Crivo-400M are
not meaningful: Crivo-150M is 0.01 points above Crivo-68M, far inside the 0.9-point rank-group rule, and Crivo-400M
precedes Amber 0.4B (25.41) by 0.04M parameters. Crivo-1B (\FullOneB) is the highest-scoring encoder on the board but
not on the frontier, since two smaller 0.8B decoders in its rank group, OneJev1.6 (29.66) and StartLux-Decision
(29.02), score higher. The field at these sizes is sparse: \NSmall{} sized entries up to 420M, \NSmallCrivo{} of them
ours.

\textbf{Where the scores come from.} Same-skill scores rise with size (15.88 to 38.75) except at 150M
(Section~\ref{sec:sel-neg}). New-domain scores stay between 3.2 and 5.5 up to 150M, below Prumo's \NDPrumo, and rise
only at 400M (9.09) and 1B (11.39): unfamiliar decision tasks are where small encoders are weakest.

\subsection{Public index and private tests}
Public-suite skill (pub15 in Section~\ref{sec:ablation}) is the primary endpoint of the reader study, so its relation to the private tests matters. Across
all \NModels{} entries of \di 0.4, the public index and the private same-skill score correlate with Pearson
$r=\RPubSS$ \RPubSSci{} and Spearman $\rho=\RhoPubSS$ \RhoPubSSci{} (the same on the mapped scale, since the mapping
is affine). It is not only an artefact of pooling 17M with 30B models: within \NBuckets{} size buckets it stays
between \RBucketMin{} and \RBucketMax, and the partial correlation after controlling for log parameters is
\RPartial{} \RPartialci{} (the bootstrap is not clustered, Section~\ref{sec:proto-stats}, so the intervals are too narrow).

\subsection{A negative result: development panels moved with the training data}
\label{sec:sel-neg}
Our internal development panels failed as predictors of the private tests. For every released model they
overestimated the private same-skill score by 5.6--7.1 points, and they ranked two training recipes in the wrong
order. The released recipe differs from the one used for the v1 models in several ways at once: it caps training on
the suite's own source splits (v1 used about 210k such rows, uncapped and in the suite's layouts), rewrites half of
those rows into other layouts, and adds sibling sources, code-generated families and a teacher. Under the new recipe
both panels rose at both sizes (same-skill panel 0.459 to 0.556 at 150M and 0.390 to 0.453 at 32M; internal numbers),
yet the board's same-skill score on identical private tests\footnote{The v1 models are scored only in the archived
edition 0.3.2, whose same-skill tests are those of edition 0.4: for the 157 entries scored in both, $S$, the public
index and all 5,338 per-benchmark private scores are identical.} fell from \SSOneFiftyVOne{} to \SSOneFifty{} at 150M
and from 18.84 to 18.44 at 32M. Per benchmark the 150M change is not significant (mean private skill \VOneTwoDiff{} \VOneTwoDiffci,
Wilcoxon $p=\VOneTwoWilcoxon$), but S1-Arena, on which neither recipe was tuned, points the same way at the item level
(\SOneVTwoMinusVOne{}, \SOneVTwoMinusVOneitem; the task interval includes zero). The public index fell with the
private score at 150M but rose at 32M (25.06 to 27.34). The panels were built from sibling sources of the board's
skills, and the new recipe added sources of exactly that kind: a development panel drawn from the same pool as new
training data is no longer an independent estimate of performance on fresh items, a close relative of the adaptive
holdout reuse described by Dwork et~al.\ \cite{dwork2015reusable}.

\subsection{What a rule engine reveals about the board}
With no learned parameters, Prumo scores a Full of \FullPrumo{} on the private tests ($S=\SSPrumo$, $O=\NDPrumo$),
above Crivo-17M (\FullSeventeen) and both Bekko models at 17M and 68M (5.72, 9.99), and below Crivo-32M
(\FullThirtyTwo). On the same skills it ties Crivo-17M (\SSPrumo{} against \SSSeventeen), and on new domains it is above
every Crivo below 400M, probably because some new-domain tasks contain structures it solves exactly. On the external
benchmarks it is near chance (S1-Arena \SOnePrumo, BTZSC \BTPrumo). The natural reading is that part of the board
rewards format-specific solving, which a hand-written engine can do without any notion of the task; this weakens the
board as evidence for the encoders, and is why we place it after the external benchmarks. Its board ECE
(\BoardEcePrumo) is low mostly because unclaimed questions receive uniform answers.

\Figure[t!](topskip=0pt, botskip=0pt, midskip=0pt)[width=0.9\columnwidth]{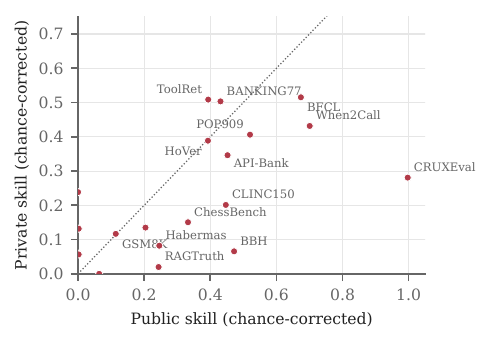}
{Prumo, public against private chance-corrected skill per benchmark (board data; dotted: $y=x$). Lexical
heuristics (retrieval, label overlap, grounding) keep their skill on fresh items; exact solvers keyed to a format
(code execution on CRUXEval, BBH solvers, chess) lose most of it.\label{fig:prumo}}

Because Prumo learns nothing, its public--private gap on a benchmark measures how well a hand-written capability
transfers to fresh items of the same skill (Fig.~\ref{fig:prumo}; $r=\RPrumoTransfer$ over \NPrumoTransfer{}
benchmarks): lexical capabilities hold (tool retrieval 0.39 public, 0.51 private), exact solvers do not (CRUXEval
0.997, 0.28; BBH 0.47, 0.07), presumably because a parser exact on the public format stops firing on another
presentation.

\section{Discussion and Limitations}
\label{sec:discussion}

\textbf{Implications.} For practitioners the reader choice reduces to two questions: is a pair-trained (reranker)
checkpoint available, and how heavy is the tail of the request distribution? With a reranker and short option lists,
per-option scoring is the more accurate design and is order-invariant by construction; with long shared states (tool
catalogues, large intent sets) or a tight mean-latency budget, the joint reader costs one pass per question and, from
a raw encoder, showed no loss in our runs. For benchmark maintainers, a parameter-free engine that matches a 17M model
on same-skill private tests is a cheap probe for format-specific shortcuts.

\textbf{Reader study.} The study uses one 150k-row subset and one size (32M); neither reader was trained with the full
recipe in a controlled way or above 32M. J-comp reaches P's compute by repeating rows, so a joint arm on about 3.25
times as many distinct rows would be the stricter control. A UniMC-style masked-attention joint reader would separate
``options do not see each other'' from ``the state is re-read per option''.

\textbf{External evaluation.} The S1-Arena audit is name- and string-based, and only four BTZSC datasets are unseen.
The comparison is with generic NLI classifiers and small LLMs, not the closest same-interface encoders
(Section~\ref{sec:proto-base}), and it mixes backbone, readout and training data; only the readout is isolated.

\textbf{Board, baselines and latency.} Private scores come from one run of each model on tests we cannot inspect. We
did not fine-tune DeBERTa-v3-large per task, the supervised ceiling reported elsewhere \cite{rafe2026benchmarking}.
Latency has no batched throughput or operation counts, and the \CpuN-request CPU sample makes means sensitive to a few
long requests.

\textbf{Use and scope.} The models are over-confident on new tasks, so thresholds for routing actions must be set on
the target task; in a hand-coded sample of 70 Crivo-68M errors on the public suite, the most confident errors were
wrong tool or action selections. They do not replace supervised in-task training or large models on knowledge-heavy
and new-domain decisions. Training and evaluation are in English only. The Crivo weights are CC BY-NC 4.0 because
several training sources are non-commercial; the Ettin bases and Prumo are Apache-2.0.

\section{Data and Code Availability}
\label{sec:availability}

All artefacts are public on the Hugging Face Hub under the account \texttt{tardellirs}:
\begin{itemize}
\item model weights and cards: \url{https://huggingface.co/tardellirs/crivo-17m}, \texttt{crivo-32m}, \texttt{crivo-68m},
\texttt{crivo-150m}, \texttt{crivo-400m} and \texttt{crivo-1b} (the v1 models are earlier revisions of \texttt{crivo-32m}
and \texttt{crivo-150m}: Crivo-32M v1 is revision \texttt{bbf3f19}, Crivo-150M v1 revision \texttt{7a6c745}); licence
CC BY-NC 4.0;
\item the three per-option 32M checkpoints of the reader study (arm P, seeds 0--2) with their inference engine:
\PerOptRepo (CC BY-NC 4.0);
\item Prumo: \url{https://huggingface.co/tardellirs/prumo} and \url{https://github.com/tardellirs/prumo} (Apache-2.0);
public-suite run outputs: the dataset \texttt{tardellirs/decision-index-results}.
\end{itemize}
The \di kit and leaderboard are public \cite{apolinario2026decisionindex,multimodalart2026jevdecisionindex}, as are
S1-Arena and BTZSC. The experiment code of this paper (training code, benchmark adapters, baseline wrappers,
contamination audit, ablation grid, statistics), the scripts that generate every table and figure from the
board snapshot and our run outputs, and the experiment log, which records the plan and decision rule of the
compute-matched and initialisation controls before they were run, are at \CodeRepo, together with an archived copy of the board
snapshot of 2026-10-20. The training mixes (Section~\ref{sec:method}) are not redistributed, since several sources carry their own licences.

\section{Conclusion}
\label{sec:conclusion}

Joint or per-option? At 32M and matched compute, the answer depends on the starting checkpoint. From a cross-encoder
reranker, whose pair input matches per-option reading, scoring each option separately is more accurate
(\AOneCompSkill{} public-suite skill, with the same sign on two external benchmarks but not significantly on their
unseen tasks); from the raw encoder, the cheaper joint reader is as good. Per-option scoring costs 5--13 times the
encoder passes, little at the median and much in the tail, so the reader should follow the application's latency
budget. How small? Joint encoders from 68M up beat a 435M zero-shot NLI classifier on all tasks of an
overlap-audited typed-decision benchmark, models up to 150M are at least on par with it on unseen tasks, and the 17M
model answers a median request in about 14~ms on a CPU. On private leaderboard tests the smallest models lead their
size range on in-distribution skills but stay near the floor on new domains, and a rule engine with no learned
parameters matches the smallest model on the same-skill tests. Future work should train the per-option reader with
the full recipe and above 32M, compare it with the closest same-interface encoders under the same audit, and widen
coverage of new domains.

\section*{Acknowledgment}
The author thanks the maintainers of the Decision Index for running the private tests, and the authors of S1-Arena
and BTZSC for releasing data, code and baseline results.

\textbf{Use of generative AI.} An AI assistant, Claude Code (Anthropic), was used to write and debug code and to
revise the text. The author verified all content and takes full responsibility for it.

{\footnotesize
\bibliographystyle{IEEEtran}
\bibliography{refs}

\begin{thebibliography}{100}
\providecommand{\url}[1]{#1}
\csname url@samestyle\endcsname
\providecommand{\newblock}{\relax}
\providecommand{\bibinfo}[2]{#2}
\providecommand{\BIBentrySTDinterwordspacing}{\spaceskip=0pt\relax}
\providecommand{\BIBentryALTinterwordstretchfactor}{4}
\providecommand{\BIBentryALTinterwordspacing}{\spaceskip=\fontdimen2\font plus
\BIBentryALTinterwordstretchfactor\fontdimen3\font minus
  \fontdimen4\font\relax}
\providecommand{\BIBforeignlanguage}[2]{{%
\expandafter\ifx\csname l@#1\endcsname\relax
\typeout{** WARNING: IEEEtran.bst: No hyphenation pattern has been}%
\typeout{** loaded for the language `#1'. Using the pattern for}%
\typeout{** the default language instead.}%
\else
\language=\csname l@#1\endcsname
\fi
#2}}
\providecommand{\BIBdecl}{\relax}
\BIBdecl

\bibitem{zheng2024robust}
\BIBentryALTinterwordspacing
C.~Zheng, H.~Zhou, F.~Meng, J.~Zhou, and M.~Huang, ``Large language models are
  not robust multiple choice selectors,'' in \emph{Proc. Int. Conf. Learning
  Representations (ICLR)}, 2024, arXiv:2309.03882. [Online]. Available:
  \url{https://arxiv.org/abs/2309.03882}
\BIBentrySTDinterwordspacing

\bibitem{pezeshkpour2024order}
P.~Pezeshkpour and E.~Hruschka, ``Large language models sensitivity to the
  order of options in multiple-choice questions,'' in \emph{Findings of the
  Association for Computational Linguistics: NAACL 2024}, 2024, pp. 2006--2017.

\bibitem{yin2019benchmarking}
W.~Yin, J.~Hay, and D.~Roth, ``Benchmarking zero-shot text classification:
  Datasets, evaluation and entailment approach,'' in \emph{Proc. 2019 Conf.
  Empirical Methods in Natural Language Processing and 9th Int. Joint Conf.
  Natural Language Processing (EMNLP-IJCNLP)}, 2019, pp. 3914--3923.

\bibitem{laurer2023building}
M.~Laurer, W.~van Atteveldt, A.~Casas, and K.~Welbers, ``Building efficient
  universal classifiers with natural language inference,'' arXiv:2312.17543,
  2023.

\bibitem{almeida2026jev}
\BIBentryALTinterwordspacing
D.~Almeida, ``Introducing {System One} models \& {Jev},'' TypeSafe AI blog,
  Sep. 2026, accessed 2026-10-20. [Online]. Available:
  \url{https://typesafe.ai/blog/introducing-system-one-models-and-jev}
\BIBentrySTDinterwordspacing

\bibitem{antunes2026ndm}
C.~A. Antunes, A.~Albuquerque, G.~de~Melo~Schmith, N.~Batista, P.~O.~E.
  da~Silva \emph{et~al.}, ``Narrow decision models: A {System-1} architecture
  for calibrated, typed decisions,'' arXiv:2610.25553, 2026.

\bibitem{li2026jeva}
Q.~Li and X.~Wang, ``Jeva: {System One} models from first principles,''
  arXiv:2610.18773, 2026.

\bibitem{devlin2019bert}
\BIBentryALTinterwordspacing
J.~Devlin, M.-W. Chang, K.~Lee, and K.~Toutanova, ``{BERT}: Pre-training of
  deep bidirectional transformers for language understanding,'' in
  \emph{Proceedings of the 2019 Conference of the {N}orth {A}merican Chapter of
  the {A}ssociation for {C}omputational {L}inguistics: Human Language
  Technologies, Volume 1 (Long and Short Papers)}.\hskip 1em plus 0.5em minus
  0.4em\relax Minneapolis, Minnesota: Association for Computational
  Linguistics, Jun. 2019, pp. 4171--4186. [Online]. Available:
  \url{https://aclanthology.org/N19-1423/}
\BIBentrySTDinterwordspacing

\bibitem{zellers2018swag}
R.~Zellers, Y.~Bisk, R.~Schwartz, and Y.~Choi, ``{SWAG}: A large-scale
  adversarial dataset for grounded commonsense inference,'' in
  \emph{Proceedings of the 2018 Conference on Empirical Methods in Natural
  Language Processing ({EMNLP})}.\hskip 1em plus 0.5em minus 0.4em\relax
  Brussels, Belgium: Association for Computational Linguistics, 2018, pp.
  93--104.

\bibitem{yang2022unimc}
P.~Yang, J.~Wang, R.~Gan, X.~Zhu, L.~Zhang, Z.~Wu, X.~Gao, J.~Zhang, and
  T.~Sakai, ``Zero-shot learners for natural language understanding via a
  unified multiple choice perspective,'' in \emph{Proceedings of the 2022
  Conference on Empirical Methods in Natural Language Processing
  ({EMNLP})}.\hskip 1em plus 0.5em minus 0.4em\relax Abu Dhabi, United Arab
  Emirates: Association for Computational Linguistics, 2022, pp. 7042--7055.

\bibitem{duarte2026dinah0}
\BIBentryALTinterwordspacing
{Lukita}, ``{Dinah-0}: A {150M}-parameter encoder that makes typed decisions,''
  Hugging Face model card, 2026, accessed 2026-10-20. [Online]. Available:
  \url{https://huggingface.co/Lukitaduarte/dinah-0}
\BIBentrySTDinterwordspacing

\bibitem{weller2025ettin}
\BIBentryALTinterwordspacing
O.~Weller, K.~Ricci, M.~Marone, A.~Chaffin, D.~Lawrie, and B.~{Van Durme},
  ``{Seq vs Seq}: An open suite of paired encoders and decoders,'' in
  \emph{Proceedings of the Fourteenth International Conference on Learning
  Representations ({ICLR})}, 2026. [Online]. Available:
  \url{https://openreview.net/forum?id=z5Mn8Rxi3l}
\BIBentrySTDinterwordspacing

\bibitem{aarsen2026ettinreranker}
\BIBentryALTinterwordspacing
T.~Aarsen, ``Introducing the {Ettin} reranker family,'' Hugging Face blog post
  and model cards, 2026, accessed 2026-10-20. [Online]. Available:
  \url{https://huggingface.co/blog/ettin-reranker}
\BIBentrySTDinterwordspacing

\bibitem{multimodalart2026jevdecisionindex}
\BIBentryALTinterwordspacing
{multimodalart}, ``{Jev} {Decision Index},'' Hugging Face Space (leaderboard),
  2026, accessed 2026-10-20. [Online]. Available:
  \url{https://huggingface.co/spaces/multimodalart/jev-decision-index}
\BIBentrySTDinterwordspacing

\bibitem{chen2026system1arena}
T.~Chen, X.~Liu, R.~An, Z.~Wang, and M.~Zhou, ``{System1-Arena}: Which
  decisions can be left to system 1?'' arXiv:2610.21244, 2026.

\bibitem{aarab2026btzsc}
\BIBentryALTinterwordspacing
I.~Aarab, ``{BTZSC}: A benchmark for zero-shot text classification across
  cross-encoders, embedding models, rerankers and {LLMs},'' in
  \emph{Proceedings of the Fourteenth International Conference on Learning
  Representations ({ICLR})}, 2026. [Online]. Available:
  \url{https://openreview.net/forum?id=IxMryAz2p3}
\BIBentrySTDinterwordspacing

\bibitem{lai2017race}
G.~Lai, Q.~Xie, H.~Liu, Y.~Yang, and E.~Hovy, ``{RACE}: Large-scale {ReAding}
  comprehension dataset from examinations,'' in \emph{Proceedings of the 2017
  Conference on Empirical Methods in Natural Language Processing
  ({EMNLP})}.\hskip 1em plus 0.5em minus 0.4em\relax Copenhagen, Denmark:
  Association for Computational Linguistics, 2017, pp. 785--794.

\bibitem{williams2018mnli}
A.~Williams, N.~Nangia, and S.~R. Bowman, ``A broad-coverage challenge corpus
  for sentence understanding through inference,'' in \emph{Proc. 2018 Conf.
  North American Chapter Assoc. Computational Linguistics: Human Language
  Technologies (NAACL-HLT)}, 2018, pp. 1112--1122.

\bibitem{lewis2020bart}
M.~Lewis, Y.~Liu, N.~Goyal, M.~Ghazvininejad, A.~Mohamed, O.~Levy, V.~Stoyanov,
  and L.~Zettlemoyer, ``{BART}: Denoising sequence-to-sequence pre-training for
  natural language generation, translation, and comprehension,'' in \emph{Proc.
  58th Annu. Meeting Assoc. Computational Linguistics (ACL)}, 2020, pp.
  7871--7880.

\bibitem{he2023debertav3}
\BIBentryALTinterwordspacing
P.~He, J.~Gao, and W.~Chen, ``{DeBERTaV3}: Improving {DeBERTa} using
  {ELECTRA}-style pre-training with gradient-disentangled embedding sharing,''
  in \emph{Proceedings of the Eleventh International Conference on Learning
  Representations ({ICLR})}, 2023. [Online]. Available:
  \url{https://openreview.net/forum?id=sE7-XhLxHA}
\BIBentrySTDinterwordspacing

\bibitem{laurer2024less}
M.~Laurer, W.~van Atteveldt, A.~Casas, and K.~Welbers, ``Less annotating, more
  classifying: Addressing the data scarcity issue of supervised machine
  learning with deep transfer learning and {BERT-NLI},'' \emph{Political
  Analysis}, vol.~32, no.~1, pp. 84--100, 2024.

\bibitem{halder2020tars}
K.~Halder, A.~Akbik, J.~Krapac, and R.~Vollgraf, ``Task-aware representation of
  sentences for generic text classification,'' in \emph{Proceedings of the 28th
  International Conference on Computational Linguistics ({COLING})}.\hskip 1em
  plus 0.5em minus 0.4em\relax Barcelona, Spain (Online): International
  Committee on Computational Linguistics, 2020, pp. 3202--3213.

\bibitem{ran2019ocn}
Q.~Ran, P.~Li, W.~Hu, and J.~Zhou, ``Option comparison network for
  multiple-choice reading comprehension,'' arXiv:1903.03033, 2019.

\bibitem{zhang2020dcmn}
S.~Zhang, H.~Zhao, Y.~Wu, Z.~Zhang, X.~Zhou, and X.~Zhou, ``{DCMN+}: Dual
  co-matching network for multi-choice reading comprehension,'' in
  \emph{Proceedings of the {AAAI} Conference on Artificial Intelligence},
  vol.~34, no.~5, 2020, pp. 9563--9570.

\bibitem{wang2023zero}
J.~Wang, P.~Yang, R.~Gan, Y.~Zhang, J.~Zhang, and T.~Sakai, ``Zero-shot
  learners for natural language understanding via a unified multiple-choice
  perspective,'' \emph{IEEE Access}, vol.~11, pp. 142\,829--142\,845, 2023.

\bibitem{zaratiana2024gliner}
U.~Zaratiana, N.~Tomeh, P.~Holat, and T.~Charnois, ``{GLiNER}: Generalist model
  for named entity recognition using bidirectional transformer,'' in
  \emph{Proceedings of the 2024 Conference of the North {A}merican Chapter of
  the Association for Computational Linguistics: Human Language Technologies
  (Volume 1: Long Papers)}.\hskip 1em plus 0.5em minus 0.4em\relax Mexico City,
  Mexico: Association for Computational Linguistics, 2024, pp. 5364--5376.

\bibitem{stepanov2025gliclass}
I.~Stepanov, M.~Shtopko, D.~Vodianytskyi, O.~Lukashov, A.~Yavorskyi, and
  M.~Yaroshenko, ``{GLiClass}: Generalist lightweight model for sequence
  classification tasks,'' arXiv:2508.07662, 2025.

\bibitem{zhang2021laco}
X.~Zhang, Q.-W. Zhang, Z.~Yan, R.~Liu, and Y.~Cao, ``Enhancing label
  correlation feedback in multi-label text classification via multi-task
  learning,'' in \emph{Findings of the Association for Computational
  Linguistics: {ACL-IJCNLP} 2021}.\hskip 1em plus 0.5em minus 0.4em\relax
  Online: Association for Computational Linguistics, 2021, pp. 1190--1200.

\bibitem{zhang2025comprehensive}
K.~Zhang, Q.~Zhang, C.-C. Wang, and J.-S.~R. Jang, ``Comprehensive study on
  zero-shot text classification using category mapping,'' \emph{IEEE Access},
  vol.~13, pp. 23\,526--23\,546, 2025.

\bibitem{schlatt2025setencoder}
F.~Schlatt, M.~Fr{\"o}be, H.~Scells, S.~Zhuang, B.~Koopman, G.~Zuccon,
  B.~Stein, M.~Potthast, and M.~Hagen, ``{Set-Encoder}: Permutation-invariant
  inter-passage attention for listwise passage re-ranking with
  cross-encoders,'' in \emph{Advances in Information Retrieval: 47th European
  Conference on Information Retrieval ({ECIR} 2025), Part {II}}, ser. Lecture
  Notes in Computer Science, vol. 15573.\hskip 1em plus 0.5em minus 0.4em\relax
  Cham: Springer, 2025, pp. 1--19.

\bibitem{robinson2023leveraging}
\BIBentryALTinterwordspacing
J.~Robinson and D.~Wingate, ``Leveraging large language models for multiple
  choice question answering,'' in \emph{Proceedings of the Eleventh
  International Conference on Learning Representations ({ICLR})}, 2023.
  [Online]. Available: \url{https://openreview.net/forum?id=yKbprarjc5B}
\BIBentrySTDinterwordspacing

\bibitem{deusser2026evaluating}
T.~Deu{\ss}er, L.~Sparrenberg, and R.~Sifa, ``Evaluating and benchmarking the
  {System One} model {Jev},'' arXiv:2609.37647, 2026.

\bibitem{duan2026general}
F.~Duan, J.~Lin, J.~Wang, J.~Xiang, J.~Wu, X.~Zhang, H.~Yan, S.~Wang, and
  Z.~Wei, ``General decision models: Benchmarking and insights beyond {Jev},''
  arXiv:2610.03935, 2026.

\bibitem{sharma2026typedbench}
R.~Sharma, A.~B. Duncan, G.~Marceau~Caron, and S.~J. Vollmer, ``{TypedBench}: A
  benchmark for calibration, framing sensitivity, and cost in {System One}
  decision models,'' arXiv:2610.11392, 2026.

\bibitem{tang2026typed}
L.~Tang and Y.~Zheng, ``Typed decision models: An early evidence audit and
  evaluation checklist,'' arXiv:2609.32160, 2026.

\bibitem{rafe2026benchmarking}
A.~Rafe and S.~Das, ``Benchmarking {System One} decision models against trained
  classifiers and language models for automated decision gates,''
  arXiv:2610.00346, 2026.

\bibitem{geuter2026releasing}
J.~Geuter, A.~Cheng, and D.~Alvarez-Melis, ``Releasing the inner {Jev}: A
  simple recipe for open decision models,'' arXiv:2610.17205, 2026.

\bibitem{li2026llmasjev}
Y.~Li and J.~Wagle, ``{LLM-as-Jev}: {LLMs} are already {Jev}-style decision
  models -- when and how to fine-tune them,'' arXiv:2610.02076, 2026.

\bibitem{zhang2026anyjev}
J.~Zhang, T.~Yang, Y.~Shi, E.~Chen, Z.~Nie \emph{et~al.}, ``{AnyJev} technical
  report,'' arXiv:2610.00831, 2026.

\bibitem{zhuang2026startlux}
X.~Zhuang, Z.~Su, J.~Pan, N.~Wong, G.~Chen, and C.-W. Kuo,
  ``{StartLux-Decision}: {AutoThink} decision models with reinforcement
  learning for calibrated deliberative decisions,'' arXiv:2610.17309, 2026.

\bibitem{deka2026retrieval}
P.~Deka, ``From retrieval to typed decisions: Calibrated {System One} models
  from biomedical sentence encoders,'' arXiv:2610.02486, 2026.

\bibitem{tateno2026bekkodataset}
\BIBentryALTinterwordspacing
Y.~Tateno, ``{Bekko} {System One} dataset v0,'' Hugging Face dataset card,
  2026, accessed 2026-10-20. [Online]. Available:
  \url{https://huggingface.co/datasets/hotchpotch/bekko-system-one-dataset-v0}
\BIBentrySTDinterwordspacing

\bibitem{tateno2026bekkomodels}
\BIBentryALTinterwordspacing
------, ``{Bekko} {System One} v0 ({17M}, {68M}, {400M}),'' Hugging Face model
  cards, 2026, accessed 2026-10-20. [Online]. Available:
  \url{https://huggingface.co/hotchpotch/bekko-system-one-v0-68m}
\BIBentrySTDinterwordspacing

\bibitem{nandakishore2026laya}
G.~Nandakishore, ``Laya as a typed probabilistic assessor: An independent
  reproduction and a preregistered study of calibration and selective
  escalation,'' arXiv:2609.33843, 2026.

\bibitem{chouhan2026verdict}
A.~S. Chouhan and J.~Perry, ``Verdict: Encode once, decide many,''
  arXiv:2610.20778, 2026.

\bibitem{nogueira2019passage}
R.~Nogueira and K.~Cho, ``Passage re-ranking with {BERT},'' arXiv:1901.04085,
  2019.

\bibitem{reimers2019sentencebert}
\BIBentryALTinterwordspacing
N.~Reimers and I.~Gurevych, ``Sentence-{BERT}: Sentence embeddings using
  {S}iamese {BERT}-networks,'' in \emph{Proceedings of the 2019 Conference on
  Empirical Methods in Natural Language Processing and the 9th International
  Joint Conference on Natural Language Processing ({EMNLP}-{IJCNLP})}.\hskip
  1em plus 0.5em minus 0.4em\relax Hong Kong, China: Association for
  Computational Linguistics, Nov. 2019, pp. 3982--3992. [Online]. Available:
  \url{https://aclanthology.org/D19-1410/}
\BIBentrySTDinterwordspacing

\bibitem{warner2025modernbert}
\BIBentryALTinterwordspacing
B.~Warner, A.~Chaffin, B.~Clavi{\'e}, O.~Weller, O.~Hallstr{\"o}m,
  S.~Taghadouini, A.~Gallagher, R.~Biswas, F.~Ladhak, T.~Aarsen, G.~T. Adams,
  J.~Howard, and I.~Poli, ``Smarter, better, faster, longer: A modern
  bidirectional encoder for fast, memory efficient, and long context finetuning
  and inference,'' in \emph{Proceedings of the 63rd Annual Meeting of the
  {A}ssociation for {C}omputational {L}inguistics (Volume 1: Long
  Papers)}.\hskip 1em plus 0.5em minus 0.4em\relax Vienna, Austria: Association
  for Computational Linguistics, Jul. 2025, pp. 2526--2547. [Online].
  Available: \url{https://aclanthology.org/2025.acl-long.127/}
\BIBentrySTDinterwordspacing

\bibitem{hinton2015distilling}
G.~Hinton, O.~Vinyals, and J.~Dean, ``Distilling the knowledge in a neural
  network,'' arXiv:1503.02531, 2015.

\bibitem{sanh2019distilbert}
V.~Sanh, L.~Debut, J.~Chaumond, and T.~Wolf, ``{DistilBERT}, a distilled
  version of {BERT}: smaller, faster, cheaper and lighter,'' arXiv:1910.01108,
  2019.

\bibitem{wang2020minilm}
\BIBentryALTinterwordspacing
W.~Wang, F.~Wei, L.~Dong, H.~Bao, N.~Yang, and M.~Zhou, ``{MiniLM}: Deep
  self-attention distillation for task-agnostic compression of pre-trained
  transformers,'' in \emph{Advances in Neural Information Processing Systems},
  vol.~33.\hskip 1em plus 0.5em minus 0.4em\relax Curran Associates, Inc.,
  2020, pp. 5776--5788. [Online]. Available:
  \url{https://proceedings.neurips.cc/paper_files/paper/2020/hash/3f5ee243547dee91fbd053c1c4a845aa-Abstract.html}
\BIBentrySTDinterwordspacing

\bibitem{yoo2026token}
E.~Yoo, J.~Kang, S.~Choi, Y.~Kim, and J.~Sim, ``Token-based task-aware
  knowledge distillation for encoder adaptation,'' \emph{IEEE Access}, vol.~14,
  pp. 99\,526--99\,540, 2026.

\bibitem{naeini2015obtaining}
M.~P. Naeini, G.~Cooper, and M.~Hauskrecht, ``Obtaining well calibrated
  probabilities using {Bayesian} binning,'' in \emph{Proc. AAAI Conf.
  Artificial Intelligence}, vol.~29, no.~1, 2015.

\bibitem{guo2017calibration}
C.~Guo, G.~Pleiss, Y.~Sun, and K.~Q. Weinberger, ``On calibration of modern
  neural networks,'' in \emph{Proceedings of the 34th International Conference
  on Machine Learning}, ser. Proceedings of Machine Learning Research,
  vol.~70.\hskip 1em plus 0.5em minus 0.4em\relax PMLR, 2017, pp. 1321--1330.

\bibitem{brier1950verification}
G.~W. Brier, ``Verification of forecasts expressed in terms of probability,''
  \emph{Monthly Weather Review}, vol.~78, no.~1, pp. 1--3, 1950.

\bibitem{geifman2017selective}
\BIBentryALTinterwordspacing
Y.~Geifman and R.~El-Yaniv, ``Selective classification for deep neural
  networks,'' in \emph{Advances in Neural Information Processing Systems},
  vol.~30.\hskip 1em plus 0.5em minus 0.4em\relax Curran Associates, Inc.,
  2017. [Online]. Available:
  \url{https://proceedings.neurips.cc/paper_files/paper/2017/hash/4a8423d5e91fda00bb7e46540e2b0cf1-Abstract.html}
\BIBentrySTDinterwordspacing

\bibitem{desai2020calibration}
S.~Desai and G.~Durrett, ``Calibration of pre-trained transformers,'' in
  \emph{Proc. 2020 Conf. Empirical Methods in Natural Language Processing
  (EMNLP)}, 2020, pp. 295--302.

\bibitem{dwork2015reusable}
C.~Dwork, V.~Feldman, M.~Hardt, T.~Pitassi, O.~Reingold, and A.~Roth, ``The
  reusable holdout: Preserving validity in adaptive data analysis,''
  \emph{Science}, vol. 349, no. 6248, pp. 636--638, 2015.

\bibitem{recht2019imagenet}
B.~Recht, R.~Roelofs, L.~Schmidt, and V.~Shankar, ``Do {ImageNet} classifiers
  generalize to {ImageNet}?'' in \emph{Proceedings of the 36th International
  Conference on Machine Learning}, ser. Proceedings of Machine Learning
  Research, vol.~97.\hskip 1em plus 0.5em minus 0.4em\relax PMLR, 2019, pp.
  5389--5400.

\bibitem{singh2025leaderboard}
S.~Singh, Y.~Nan, A.~Wang, D.~Dsouza, S.~Kapoor, A.~{\"U}st{\"u}n, S.~Koyejo,
  Y.~Deng, S.~Longpre, N.~Smith, B.~Ermis, M.~Fadaee, and S.~Hooker, ``The
  leaderboard illusion,'' in \emph{Advances in Neural Information Processing
  Systems (Datasets and Benchmarks Track)}, vol.~38.\hskip 1em plus 0.5em minus
  0.4em\relax Curran Associates, Inc., 2025, pp. 86\,910--86\,964.

\bibitem{sainz2023nlp}
\BIBentryALTinterwordspacing
O.~Sainz, J.~Campos, I.~Garc{\'i}a-Ferrero, J.~Etxaniz, O.~L. de~Lacalle, and
  E.~Agirre, ``{NLP} evaluation in trouble: On the need to measure {LLM} data
  contamination for each benchmark,'' in \emph{Findings of the {A}ssociation
  for {C}omputational {L}inguistics: {EMNLP} 2023}.\hskip 1em plus 0.5em minus
  0.4em\relax Singapore: Association for Computational Linguistics, Dec. 2023,
  pp. 10\,776--10\,787. [Online]. Available:
  \url{https://aclanthology.org/2023.findings-emnlp.722/}
\BIBentrySTDinterwordspacing

\bibitem{schwartz2020greenai}
R.~Schwartz, J.~Dodge, N.~A. Smith, and O.~Etzioni, ``Green {AI},''
  \emph{Communications of the ACM}, vol.~63, no.~12, pp. 54--63, 2020.

\bibitem{jiang2023lowresource}
Z.~Jiang, M.~Yang, M.~Tsirlin, R.~Tang, Y.~Dai, and J.~Lin, ``{`Low-Resource'}
  text classification: A parameter-free classification method with
  compressors,'' in \emph{Findings of the Association for Computational
  Linguistics: ACL 2023}, 2023, pp. 6810--6828.

\bibitem{lim2026when}
S.~Lim, H.-C. Kwon, and M.~Kim, ``When to invoke an {LLM} in industrial
  natural-language-to-{DSL} conversion: A taxonomy-driven, confidence-gated
  selective hybrid for safety-critical avionics test-script authoring,''
  \emph{IEEE Access}, vol.~14, pp. 137\,408--137\,433, 2026.

\bibitem{robertson2009probabilistic}
S.~Robertson and H.~Zaragoza, ``The probabilistic relevance framework: {BM25}
  and beyond,'' \emph{Foundations and Trends in Information Retrieval}, vol.~3,
  no.~4, pp. 333--389, 2009.

\bibitem{laurer2024zeroshotv2}
\BIBentryALTinterwordspacing
M.~Laurer, ``deberta-v3-large-zeroshot-v2.0 ({Zeroshot Classifiers v2.0}),''
  Hugging Face model card, 2024, accessed 2026-10-21. [Online]. Available:
  \url{https://huggingface.co/MoritzLaurer/deberta-v3-large-zeroshot-v2.0}
\BIBentrySTDinterwordspacing

\bibitem{yang2025qwen3}
A.~Yang, A.~Li, B.~Yang, B.~Zhang, B.~Hui \emph{et~al.}, ``{Qwen3} technical
  report,'' arXiv:2505.09388, 2025.

\bibitem{gemmateam2025gemma3}
{Gemma Team}, A.~Kamath, J.~Ferret, S.~Pathak, N.~Vieillard \emph{et~al.},
  ``{Gemma 3} technical report,'' arXiv:2503.19786, 2025.

\bibitem{apolinario2026decisionindex}
\BIBentryALTinterwordspacing
{apolinario}, ``{Decision Index}: reproduce the {Jev} decision-model benchmark
  suite locally or as one {HF Job},'' GitHub repository, 2026, accessed
  2026-10-20. [Online]. Available:
  \url{https://github.com/apolinario/decision-index}
\BIBentrySTDinterwordspacing

\bibitem{patil2025bfcl}
\BIBentryALTinterwordspacing
S.~G. Patil, H.~Mao, F.~Yan, C.~C.-J. Ji, V.~Suresh, I.~Stoica, and J.~E.
  Gonzalez, ``The {Berkeley} function calling leaderboard ({BFCL}): From tool
  use to agentic evaluation of large language models,'' in \emph{Proceedings of
  the 42nd International Conference on Machine Learning}, ser. Proceedings of
  Machine Learning Research, vol. 267.\hskip 1em plus 0.5em minus 0.4em\relax
  PMLR, 2025, pp. 48\,371--48\,392. [Online]. Available:
  \url{https://proceedings.mlr.press/v267/patil25a.html}
\BIBentrySTDinterwordspacing

\bibitem{ross2025when2call}
\BIBentryALTinterwordspacing
H.~Ross, A.~S. Mahabaleshwarkar, and Y.~Suhara, ``{W}hen2{C}all: When (not) to
  call tools,'' in \emph{Proceedings of the 2025 Conference of the Nations of
  the {A}mericas Chapter of the {A}ssociation for {C}omputational
  {L}inguistics: Human Language Technologies (Volume 1: Long Papers)}.\hskip
  1em plus 0.5em minus 0.4em\relax Albuquerque, New Mexico: Association for
  Computational Linguistics, Apr. 2025, pp. 3391--3409. [Online]. Available:
  \url{https://aclanthology.org/2025.naacl-long.174/}
\BIBentrySTDinterwordspacing

\bibitem{li2023apibank}
\BIBentryALTinterwordspacing
M.~Li, Y.~Zhao, B.~Yu, F.~Song, H.~Li, H.~Yu, Z.~Li, F.~Huang, and Y.~Li,
  ``{API}-bank: A comprehensive benchmark for tool-augmented {LLM}s,'' in
  \emph{Proceedings of the 2023 Conference on Empirical Methods in Natural
  Language Processing}.\hskip 1em plus 0.5em minus 0.4em\relax Singapore:
  Association for Computational Linguistics, Dec. 2023, pp. 3102--3116.
  [Online]. Available: \url{https://aclanthology.org/2023.emnlp-main.187/}
\BIBentrySTDinterwordspacing

\bibitem{shi2025toolret}
\BIBentryALTinterwordspacing
Z.~Shi, Y.~Wang, L.~Yan, P.~Ren, S.~Wang, D.~Yin, and Z.~Ren, ``Retrieval
  models aren{'}t tool-savvy: Benchmarking tool retrieval for large language
  models,'' in \emph{Findings of the {A}ssociation for {C}omputational
  {L}inguistics: {ACL} 2025}.\hskip 1em plus 0.5em minus 0.4em\relax Vienna,
  Austria: Association for Computational Linguistics, Jul. 2025, pp.
  24\,497--24\,524. [Online]. Available:
  \url{https://aclanthology.org/2025.findings-acl.1258/}
\BIBentrySTDinterwordspacing

\bibitem{casanueva2020banking77}
\BIBentryALTinterwordspacing
I.~Casanueva, T.~Tem{\v{c}}inas, D.~Gerz, M.~Henderson, and I.~Vuli{\'c},
  ``Efficient intent detection with dual sentence encoders,'' in
  \emph{Proceedings of the 2nd Workshop on Natural Language Processing for
  Conversational {AI}}.\hskip 1em plus 0.5em minus 0.4em\relax Online:
  Association for Computational Linguistics, Jul. 2020, pp. 38--45. [Online].
  Available: \url{https://aclanthology.org/2020.nlp4convai-1.5/}
\BIBentrySTDinterwordspacing

\bibitem{larson2019clinc}
\BIBentryALTinterwordspacing
S.~Larson, A.~Mahendran, J.~J. Peper, C.~Clarke, A.~Lee, P.~Hill, J.~K.
  Kummerfeld, K.~Leach, M.~A. Laurenzano, L.~Tang, and J.~Mars, ``An evaluation
  dataset for intent classification and out-of-scope prediction,'' in
  \emph{Proceedings of the 2019 Conference on Empirical Methods in Natural
  Language Processing and the 9th International Joint Conference on Natural
  Language Processing ({EMNLP}-{IJCNLP})}.\hskip 1em plus 0.5em minus
  0.4em\relax Hong Kong, China: Association for Computational Linguistics, Nov.
  2019, pp. 1311--1316. [Online]. Available:
  \url{https://aclanthology.org/D19-1131/}
\BIBentrySTDinterwordspacing

\bibitem{jiang2020hover}
\BIBentryALTinterwordspacing
Y.~Jiang, S.~Bordia, Z.~Zhong, C.~Dognin, M.~Singh, and M.~Bansal, ``{H}o{V}er:
  A dataset for many-hop fact extraction and claim verification,'' in
  \emph{Findings of the {A}ssociation for {C}omputational {L}inguistics:
  {EMNLP} 2020}.\hskip 1em plus 0.5em minus 0.4em\relax Online: Association for
  Computational Linguistics, Nov. 2020, pp. 3441--3460. [Online]. Available:
  \url{https://aclanthology.org/2020.findings-emnlp.309/}
\BIBentrySTDinterwordspacing

\bibitem{niu2024ragtruth}
\BIBentryALTinterwordspacing
C.~Niu, Y.~Wu, J.~Zhu, S.~Xu, K.~Shum, R.~Zhong, J.~Song, and T.~Zhang,
  ``{RAGT}ruth: A hallucination corpus for developing trustworthy
  retrieval-augmented language models,'' in \emph{Proceedings of the 62nd
  Annual Meeting of the {A}ssociation for {C}omputational {L}inguistics (Volume
  1: Long Papers)}.\hskip 1em plus 0.5em minus 0.4em\relax Bangkok, Thailand:
  Association for Computational Linguistics, Aug. 2024, pp. 10\,862--10\,878.
  [Online]. Available: \url{https://aclanthology.org/2024.acl-long.585/}
\BIBentrySTDinterwordspacing

\bibitem{nie2020anli}
\BIBentryALTinterwordspacing
Y.~Nie, A.~Williams, E.~Dinan, M.~Bansal, J.~Weston, and D.~Kiela,
  ``Adversarial {NLI}: A new benchmark for natural language understanding,'' in
  \emph{Proceedings of the 58th Annual Meeting of the {A}ssociation for
  {C}omputational {L}inguistics}.\hskip 1em plus 0.5em minus 0.4em\relax
  Online: Association for Computational Linguistics, Jul. 2020, pp. 4885--4901.
  [Online]. Available: \url{https://aclanthology.org/2020.acl-main.441/}
\BIBentrySTDinterwordspacing

\bibitem{koreeda2021contractnli}
\BIBentryALTinterwordspacing
Y.~Koreeda and C.~Manning, ``{C}ontract{NLI}: A dataset for document-level
  natural language inference for contracts,'' in \emph{Findings of the
  {A}ssociation for {C}omputational {L}inguistics: {EMNLP} 2021}.\hskip 1em
  plus 0.5em minus 0.4em\relax Punta Cana, Dominican Republic: Association for
  Computational Linguistics, Nov. 2021, pp. 1907--1919. [Online]. Available:
  \url{https://aclanthology.org/2021.findings-emnlp.164/}
\BIBentrySTDinterwordspacing

\bibitem{jullien2024nli4ct}
\BIBentryALTinterwordspacing
M.~Jullien, M.~Valentino, and A.~Freitas, ``{S}em{E}val-2024 task 2: Safe
  biomedical natural language inference for clinical trials,'' in
  \emph{Proceedings of the 18th {I}nternational {W}orkshop on {S}emantic
  {E}valuation ({S}em{E}val-2024)}.\hskip 1em plus 0.5em minus 0.4em\relax
  Mexico City, Mexico: Association for Computational Linguistics, Jun. 2024,
  pp. 1947--1962. [Online]. Available:
  \url{https://aclanthology.org/2024.semeval-1.271/}
\BIBentrySTDinterwordspacing

\bibitem{cai2021acos}
\BIBentryALTinterwordspacing
H.~Cai, R.~Xia, and J.~Yu, ``Aspect-category-opinion-sentiment quadruple
  extraction with implicit aspects and opinions,'' in \emph{Proceedings of the
  59th Annual Meeting of the {A}ssociation for {C}omputational {L}inguistics
  and the 11th International Joint Conference on Natural Language Processing
  (Volume 1: Long Papers)}.\hskip 1em plus 0.5em minus 0.4em\relax Online:
  Association for Computational Linguistics, Aug. 2021, pp. 340--350. [Online].
  Available: \url{https://aclanthology.org/2021.acl-long.29/}
\BIBentrySTDinterwordspacing

\bibitem{tang2023finentity}
\BIBentryALTinterwordspacing
Y.~Tang, Y.~Yang, A.~Huang, A.~Tam, and J.~Tang, ``{F}in{E}ntity: Entity-level
  sentiment classification for financial texts,'' in \emph{Proceedings of the
  2023 Conference on Empirical Methods in Natural Language Processing}.\hskip
  1em plus 0.5em minus 0.4em\relax Singapore: Association for Computational
  Linguistics, Dec. 2023, pp. 15\,465--15\,471. [Online]. Available:
  \url{https://aclanthology.org/2023.emnlp-main.956/}
\BIBentrySTDinterwordspacing

\bibitem{allaway2020vast}
\BIBentryALTinterwordspacing
E.~Allaway and K.~McKeown, ``{Z}ero-{S}hot {S}tance {D}etection: {A} {D}ataset
  and {M}odel using {G}eneralized {T}opic {R}epresentations,'' in
  \emph{Proceedings of the 2020 Conference on Empirical Methods in Natural
  Language Processing ({EMNLP})}.\hskip 1em plus 0.5em minus 0.4em\relax
  Online: Association for Computational Linguistics, Nov. 2020, pp. 8913--8931.
  [Online]. Available: \url{https://aclanthology.org/2020.emnlp-main.717/}
\BIBentrySTDinterwordspacing

\bibitem{abufarha2022isarcasmeval}
\BIBentryALTinterwordspacing
I.~Abu~Farha, S.~V. Oprea, S.~Wilson, and W.~Magdy, ``{S}em{E}val-2022 task 6:
  i{S}arcasm{E}val, intended sarcasm detection in {E}nglish and {A}rabic,'' in
  \emph{Proceedings of the 16th {I}nternational {W}orkshop on {S}emantic
  {E}valuation ({S}em{E}val-2022)}.\hskip 1em plus 0.5em minus 0.4em\relax
  Seattle, United States: Association for Computational Linguistics, Jul. 2022,
  pp. 802--814. [Online]. Available:
  \url{https://aclanthology.org/2022.semeval-1.111/}
\BIBentrySTDinterwordspacing

\bibitem{zellers2019hellaswag}
\BIBentryALTinterwordspacing
R.~Zellers, A.~Holtzman, Y.~Bisk, A.~Farhadi, and Y.~Choi, ``{H}ella{S}wag: Can
  a machine really finish your sentence?'' in \emph{Proceedings of the 57th
  Annual Meeting of the {A}ssociation for {C}omputational {L}inguistics}.\hskip
  1em plus 0.5em minus 0.4em\relax Florence, Italy: Association for
  Computational Linguistics, Jul. 2019, pp. 4791--4800. [Online]. Available:
  \url{https://aclanthology.org/P19-1472/}
\BIBentrySTDinterwordspacing

\bibitem{sakaguchi2020winogrande}
K.~Sakaguchi, R.~Le~Bras, C.~Bhagavatula, and Y.~Choi, ``{WinoGrande}: An
  adversarial {Winograd} schema challenge at scale,'' in \emph{Proceedings of
  the AAAI Conference on Artificial Intelligence}, vol.~34, 2020, pp.
  8732--8740.

\bibitem{wang2024mmlupro}
Y.~Wang, X.~Ma, G.~Zhang, Y.~Ni, A.~Chandra, S.~Guo, W.~Ren, A.~Arulraj, X.~He,
  Z.~Jiang, T.~Li, M.~Ku, K.~Wang, A.~Zhuang, R.~Fan, X.~Yue, and W.~Chen,
  ``{MMLU-Pro}: A more robust and challenging multi-task language understanding
  benchmark,'' in \emph{Advances in Neural Information Processing Systems},
  vol.~37, 2024, pp. 95\,266--95\,290.

\bibitem{rein2023gpqa}
\BIBentryALTinterwordspacing
D.~Rein, B.~L. Hou, A.~C. Stickland, J.~Petty, R.~Y. Pang, J.~Dirani,
  J.~Michael, and S.~R. Bowman, ``{GPQA}: A graduate-level {Google}-proof
  {Q\&A} benchmark,'' in \emph{Proceedings of the First Conference on Language
  Modeling ({COLM})}, 2024. [Online]. Available:
  \url{https://openreview.net/forum?id=Ti67584b98}
\BIBentrySTDinterwordspacing

\bibitem{phan2025hle}
{Center for AI Safety}, {Scale AI}, and {HLE Contributors Consortium}, ``A
  benchmark of expert-level academic questions to assess {AI} capabilities,''
  \emph{Nature}, vol. 649, no. 8099, pp. 1139--1146, 2026.

\bibitem{suzgun2023bbh}
\BIBentryALTinterwordspacing
M.~Suzgun, N.~Scales, N.~Sch{\"a}rli, S.~Gehrmann, Y.~Tay, H.~W. Chung,
  A.~Chowdhery, Q.~Le, E.~Chi, D.~Zhou, and J.~Wei, ``Challenging {BIG}-bench
  tasks and whether chain-of-thought can solve them,'' in \emph{Findings of the
  {A}ssociation for {C}omputational {L}inguistics: {ACL} 2023}.\hskip 1em plus
  0.5em minus 0.4em\relax Toronto, Canada: Association for Computational
  Linguistics, Jul. 2023, pp. 13\,003--13\,051. [Online]. Available:
  \url{https://aclanthology.org/2023.findings-acl.824/}
\BIBentrySTDinterwordspacing

\bibitem{cobbe2021gsm8k}
K.~Cobbe, V.~Kosaraju, M.~Bavarian, M.~Chen, H.~Jun, L.~Kaiser, M.~Plappert,
  J.~Tworek, J.~Hilton, R.~Nakano, C.~Hesse, and J.~Schulman, ``Training
  verifiers to solve math word problems,'' arXiv:2110.14168, 2021.

\bibitem{gu2024cruxeval}
A.~Gu, B.~Rozière, H.~Leather, A.~Solar-Lezama, G.~Synnaeve, and S.~I. Wang,
  ``{CRUXEval}: A benchmark for code reasoning, understanding and execution,''
  in \emph{Proceedings of the 41st International Conference on Machine
  Learning}, ser. Proceedings of Machine Learning Research, vol. 235.\hskip 1em
  plus 0.5em minus 0.4em\relax PMLR, 2024, pp. 16\,568--16\,621.

\bibitem{jin2023cladder}
Z.~Jin, Y.~Chen, F.~Leeb, L.~Gresele, O.~Kamal, Z.~Lyu, K.~Blin, F.~G. Adauto,
  M.~Kleiman-Weiner, M.~Sachan, and B.~Schölkopf, ``{CLadder}: Assessing
  causal reasoning in language models,'' in \emph{Advances in Neural
  Information Processing Systems}, vol.~36, 2023, pp. 31\,038--31\,065.

\bibitem{sprague2024musr}
Z.~Sprague, X.~Ye, K.~Bostrom, S.~Chaudhuri, and G.~Durrett, ``{MuSR}: Testing
  the limits of chain-of-thought with multistep soft reasoning,'' in \emph{The
  Twelfth International Conference on Learning Representations (ICLR)}, 2024,
  pp. 14\,670--14\,728.

\bibitem{ruoss2024chessbench}
A.~Ruoss, G.~Delétang, S.~Medapati, J.~Grau-Moya, L.~K. Wenliang, E.~Catt,
  J.~Reid, C.~A. Lewis, J.~Veness, and T.~Genewein, ``Amortized planning with
  large-scale transformers: A case study on chess,'' in \emph{Advances in
  Neural Information Processing Systems}, vol.~37, 2024, pp. 65\,765--65\,790.

\bibitem{reddy2022esci}
C.~K. Reddy, L.~Màrquez, F.~Valero, N.~Rao, H.~Zaragoza, S.~Bandyopadhyay,
  A.~Biswas, A.~Xing, and K.~Subbian, ``Shopping queries dataset: A
  {Large-Scale} {ESCI} benchmark for improving product search,''
  arXiv:2206.06588, 2022.

\bibitem{su2024bright}
\BIBentryALTinterwordspacing
H.~Su, H.~Yen, M.~Xia, W.~Shi, N.~Muennighoff, H.~Wang, H.~Liu, Q.~Shi, Z.~S.
  Siegel, M.~Tang, R.~Sun, J.~Yoon, S.~O. Arik, D.~Chen, and T.~Yu, ``{BRIGHT}:
  A realistic and challenging benchmark for reasoning-intensive retrieval,'' in
  \emph{Proceedings of the Thirteenth International Conference on Learning
  Representations ({ICLR})}, 2025, pp. 48\,941--48\,991. [Online]. Available:
  \url{https://openreview.net/forum?id=ykuc5q381b}
\BIBentrySTDinterwordspacing

\bibitem{hossain2019humicroedit}
\BIBentryALTinterwordspacing
N.~Hossain, J.~Krumm, and M.~Gamon, ```{P}resident vows to cut
  {\ensuremath{<}}{T}axes{\ensuremath{>}} {H}air': Dataset and analysis of
  creative text editing for humorous headlines,'' in \emph{Proceedings of the
  2019 Conference of the {N}orth {A}merican Chapter of the {A}ssociation for
  {C}omputational {L}inguistics: Human Language Technologies, Volume 1 (Long
  and Short Papers)}.\hskip 1em plus 0.5em minus 0.4em\relax Minneapolis,
  Minnesota: Association for Computational Linguistics, Jun. 2019, pp.
  133--142. [Online]. Available: \url{https://aclanthology.org/N19-1012/}
\BIBentrySTDinterwordspacing

\bibitem{hessel2023androids}
\BIBentryALTinterwordspacing
J.~Hessel, A.~Marasovic, J.~D. Hwang, L.~Lee, J.~Da, R.~Zellers, R.~Mankoff,
  and Y.~Choi, ``Do androids laugh at electric sheep? {H}umor `understanding'
  benchmarks from {The New Yorker Caption Contest},'' in \emph{Proceedings of
  the 61st Annual Meeting of the {A}ssociation for {C}omputational
  {L}inguistics (Volume 1: Long Papers)}.\hskip 1em plus 0.5em minus
  0.4em\relax Toronto, Canada: Association for Computational Linguistics, Jul.
  2023, pp. 688--714. [Online]. Available:
  \url{https://aclanthology.org/2023.acl-long.41/}
\BIBentrySTDinterwordspacing

\bibitem{yao2025bachi}
M.~Yao, K.~Chen, S.~Dubnov, and T.~Berg-Kirkpatrick, ``{BACHI}: Boundary-aware
  symbolic chord recognition through masked iterative decoding on pop and
  classical music,'' in \emph{Proceedings of the IEEE International Conference
  on Acoustics, Speech and Signal Processing ({ICASSP})}, Barcelona, Spain,
  2026, pp. 15\,257--15\,261.

\bibitem{tessler2024habermas}
M.~H. Tessler, M.~A. Bakker, D.~Jarrett, H.~Sheahan, M.~J. Chadwick, R.~Koster,
  G.~Evans, L.~Campbell-Gillingham, T.~Collins, D.~C. Parkes, M.~Botvinick, and
  C.~Summerfield, ``{AI} can help humans find common ground in democratic
  deliberation,'' \emph{Science}, vol. 386, no. 6719, p. eadq2852, 2024.

\bibitem{sileo2024tasksource}
\BIBentryALTinterwordspacing
D.~Sileo, ``tasksource: A large collection of {NLP} tasks with a structured
  dataset preprocessing framework,'' in \emph{Proceedings of the 2024 Joint
  International Conference on Computational Linguistics, Language Resources and
  Evaluation ({LREC}-{COLING} 2024)}.\hskip 1em plus 0.5em minus 0.4em\relax
  Torino, Italia: ELRA and ICCL, May 2024, pp. 15\,655--15\,684. [Online].
  Available: \url{https://aclanthology.org/2024.lrec-main.1361/}
\BIBentrySTDinterwordspacing

\bibitem{tasksource2026jevtyped}
\BIBentryALTinterwordspacing
{tasksource}, ``{tasksource-jev-typed-decisions},'' Hugging Face dataset card,
  2026, accessed 2026-10-20. [Online]. Available:
  \url{https://huggingface.co/datasets/tasksource/tasksource-jev-typed-decisions}
\BIBentrySTDinterwordspacing

\bibitem{tasksource2026procedural}
\BIBentryALTinterwordspacing
------, ``{procedural-typed-decisions},'' Hugging Face dataset card, 2026,
  accessed 2026-10-20. [Online]. Available:
  \url{https://huggingface.co/datasets/tasksource/procedural-typed-decisions}
\BIBentrySTDinterwordspacing

\bibitem{sileo2026reasoningcore}
D.~Sileo, V.~Lacombe, and D.~Kachler, ``{Reasoning Core}: Designing broad
  procedural data for {Completion-Supervised} reasoning training,''
  arXiv:2608.05148, 2026.

\bibitem{tarvainen2017meanteachers}
\BIBentryALTinterwordspacing
A.~Tarvainen and H.~Valpola, ``Mean teachers are better role models:
  Weight-averaged consistency targets improve semi-supervised deep learning
  results,'' in \emph{Advances in Neural Information Processing Systems},
  vol.~30.\hskip 1em plus 0.5em minus 0.4em\relax Curran Associates, Inc.,
  2017. [Online]. Available:
  \url{https://proceedings.neurips.cc/paper_files/paper/2017/hash/68053af2923e00204c3ca7c6a3150cf7-Abstract.html}
\BIBentrySTDinterwordspacing

\bibitem{wortsman2022wiseft}
M.~Wortsman, G.~Ilharco, J.~W. Kim, M.~Li, S.~Kornblith, R.~Roelofs,
  R.~Gontijo-Lopes, H.~Hajishirzi, A.~Farhadi, H.~Namkoong, and L.~Schmidt,
  ``Robust fine-tuning of zero-shot models,'' in \emph{2022 IEEE/CVF Conference
  on Computer Vision and Pattern Recognition (CVPR)}.\hskip 1em plus 0.5em
  minus 0.4em\relax IEEE, 2022, pp. 7949--7961.

\end{thebibliography}
}

\end{document}